\documentclass[11pt]{article}
\usepackage{acl}

\usepackage[utf8]{inputenc} 
\usepackage[T1]{fontenc}    
\usepackage{hyperref}       
\usepackage{url}            
\usepackage{booktabs}       
\usepackage{amsfonts}       
\usepackage{nicefrac}       
\usepackage{microtype}      
\usepackage{xcolor} 
\usepackage{colortbl}
\usepackage{natbib}
\usepackage{tikz}
\usepackage{rotating}

\usepackage{subfigure}
\usepackage{graphicx}
\usepackage{amsmath}
\usepackage{caption}
\usepackage{mathtools}
\usepackage{bm}
\usepackage{soul}
\usepackage{wrapfig}
\usepackage{algorithm}
\usepackage{algpseudocode}
\usepackage[inline]{enumitem}
\usepackage{times}
\usepackage{latexsym}
\usepackage{inconsolata}
\usepackage{amsthm}

\usepackage{titlesec}

\usepackage{makecell, multirow, threeparttable}

\definecolor{Gray}{gray}{0.9}

\newcommand{\para}[1]{\vspace{.05in}\noindent\textbf{#1}}
\newcommand{\revise}[1]{\textcolor{black}{#1}}
\newcommand{\modelname}{\texttt{CTPD}}

\definecolor{darkgreen}{rgb}{0,0.5,0}
\newcommand{\highlightgain}[1]{\textbf{\textcolor{darkgreen}{#1}}}

\title{
\modelname: Cross-Modal Temporal Pattern Discovery for Enhanced \\
Multimodal Electronic Health Records Analysis
}

\author{
 \textbf{Fuying Wang\textsuperscript{1}}$^{*}$,
 \textbf{Feng Wu\textsuperscript{1}}\thanks{Equal contributions.},
 \textbf{Yihan Tang\textsuperscript{1}},
 \textbf{Lequan Yu\textsuperscript{1}\thanks{Corresponding Author.}}
\\
\\
 \textsuperscript{1}Department of Statistics and Actuarial Science, \\
 School of Computing and Data Science, \\
 The University of Hong Kong
\\
\texttt{\{fuyingw,fengwu96,hank-tang\}@connect.hku.hk, lqyu@hku.hk}
\\
 \small{
   \textbf{Correspondence:} \href{lqyu@hku.hk}{lqyu@hku.hk}
 }
}

\begin{document}
\maketitle

\begin{abstract}

Integrating multimodal Electronic Health Records (EHR) data—such as numerical time series and free-text clinical reports—has great potential in predicting clinical outcomes.
However, prior work has primarily focused on capturing temporal interactions within individual samples and fusing multimodal information, overlooking critical temporal patterns across patients.
These patterns, such as trends in vital signs like abnormal heart rate or blood pressure, can indicate deteriorating health or an impending critical event.  Similarly, clinical notes often contain textual descriptions that reflect these patterns. Identifying corresponding temporal patterns across different modalities is crucial for improving the accuracy of clinical outcome predictions, yet it remains a challenging task.
To address this gap, we introduce a \textbf{C}ross-Modal \textbf{T}emporal \textbf{P}attern \textbf{D}iscovery (\modelname) framework, designed to efficiently extract meaningful cross-modal temporal patterns from multimodal EHR data. Our approach introduces shared initial temporal pattern representations which are refined using slot attention to generate temporal semantic embeddings. 
\revise{
To ensure rich cross-modal temporal semantics in the learned patterns, we introduce a Temporal Pattern Noise Contrastive Estimation (TP-NCE) loss for cross-modal alignment, along with two reconstruction losses to retain core information of each modality.
}
Evaluations on two clinically critical tasks—48-hour in-hospital mortality and 24-hour phenotype classification—using the MIMIC-III database demonstrate the superiority of our method over existing approaches.
The code is available at \url{https://github.com/HKU-MedAI/CTPD}.
\end{abstract}

%
\section{Introduction}
%
%
The increasing availability of Electronic Health Records (EHR) presents significant opportunities for advancing predictive modeling in healthcare~\citep{acosta2022, wang2024multimodal}. 
EHR data is inherently multimodal and time-aware, encompassing structured data like vital signs, laboratory results, and medications, as well as unstructured data such as free-text clinical reports~\cite{kim2023heterogeneous}.
Integrating these diverse data types is crucial for comprehensive patient monitoring and accurate prediction of clinical outcomes~\cite{hayat2022medfuse,wang2023hierarchical,zhang2023improving}. 
However, the irregularity and heterogeneity of multimodal data present significant challenges for precise outcome prediction.

\definecolor{our_blue}{HTML}{0F9ED5}
\definecolor{our_green}{HTML}{4EA72E}
\definecolor{our_yellow}{HTML}{FFC000}

\begin{figure*}[t!]
    \centering
    \includegraphics[width=0.9\textwidth]{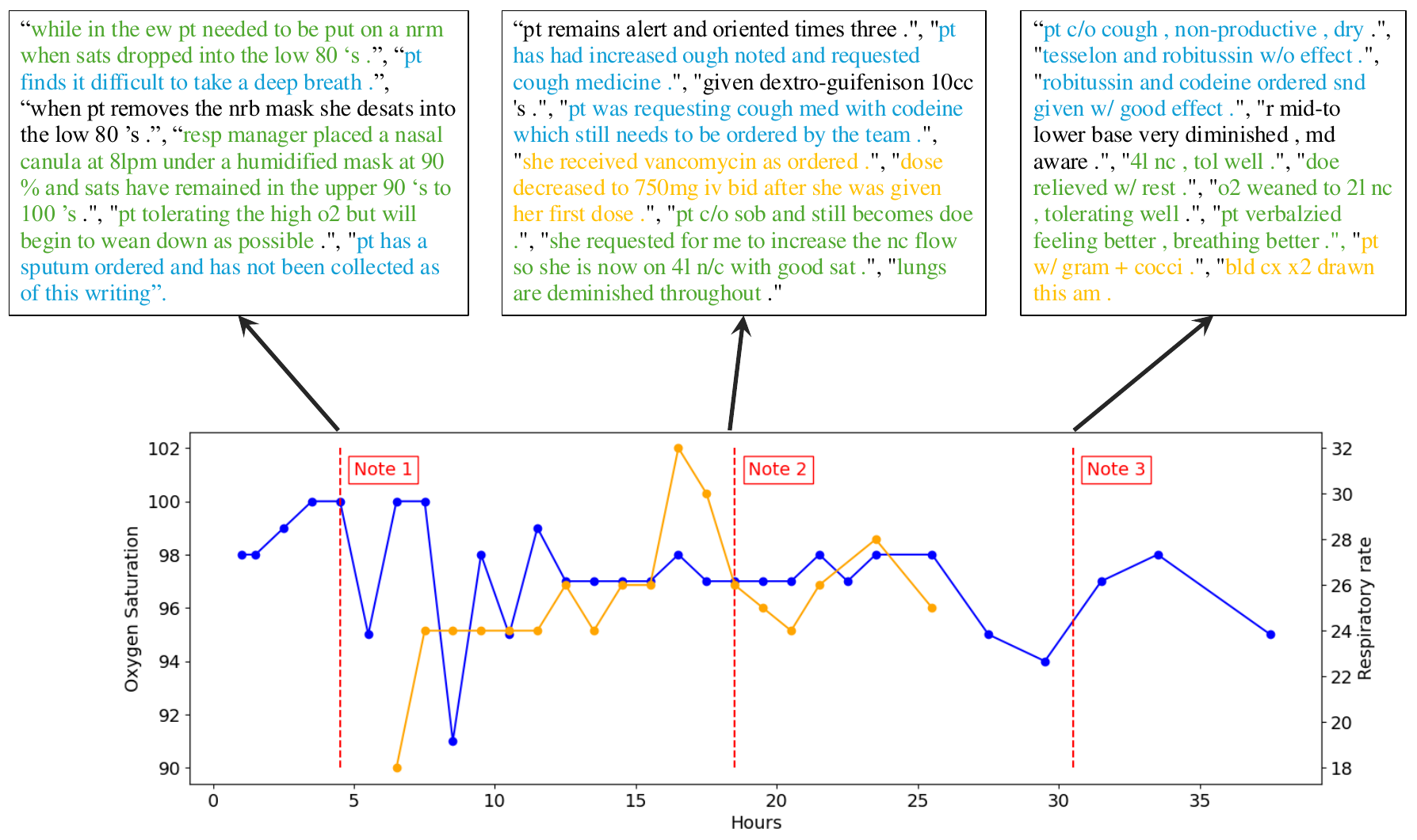}
    \caption{
    \textbf{Motivation of our proposed \modelname}: we visualized the time-series EHR with corresponding clinical notes in one ICU stay of the MIMIC-III dataset, and observed the temporal patterns across two modalities: 
    \textbf{\textcolor{our_blue}{Blue text}} highlights respiratory status. Oxygen requirements gradually decreased from 8L to 4L, and then to 2L nasal cannula, indicating steady respiratory improvement.
    \textcolor{red}{
    Note that this pattern is also reflected from the time series.
    }
    \textbf{\textcolor{our_green}{Green text}} captures cough progression and medication effects. Symptom relief was observed after administering Robitussin with codeine, demonstrating a delayed but positive response to treatment.
    \textbf{\textcolor{our_yellow}{Yellow text}} represents infection monitoring. The detection of Gram-positive cocci prompted blood culture collection (bld cx) for further evaluation, indicating active infection surveillance.
    }
    \vspace{-0.4cm}
    \label{fig:teaser}
\end{figure*}
%

Existing approaches primarily address either the irregularity of data~\cite{shukla2019interpolation,horn2020set,zhang2021graph,zhang2023improving} or the fusion of multiple modalities~\cite{huang2020multimodal,zhang2020advances, xu2021mufasa, kline2022}, but they often neglect broader temporal trends that span across patient cases.
These cross-modal temporal patterns, present in both structured and unstructured data, can provide high-level semantic insights into a patient's health trajectory and potential risks~\cite{conrad2018temporal}.
\revise{
As illustrated in Fig.~\ref{fig:teaser}, high-level semantic patterns related to medical conditions can emerge across multiple modalities.
Capturing these patterns in a cross-modal, temporal manner is essential for improving predictive performance. 
Furthermore, critical temporal patterns in EHR data unfold at different time scales~\cite{zhang2023warpformer,luo2020hitanet,ma2020adacare,ye2020lsan}, yet existing methods struggle to capture variations across these granularities.
For instance, sudden changes in vital signs—such as a sharp drop in oxygen saturation or a rapid heart rate increase—may indicate an acute health crisis. In contrast, longer-term trends, such as persistently high blood pressure or a gradual decline in respiratory function, may signal deteriorating health or an impending critical event.
Effectively analyzing patient states across multiple time scales is crucial for comprehensive EHR modeling, yet remains an open challenge in current methodologies.
}

\revise{
To address these limitations, we propose the \textbf{C}ross-modal \textbf{T}emporal \textbf{P}attern \textbf{D}iscovery (\modelname) framework, designed to extract meaningful temporal patterns from multimodal EHR data to improve the accuracy of clinical outcome predictions.
The core innovation of our approach is a novel temporal pattern discovery module, which identifies corresponding temporal patterns (i.e., temporal prototypes) with meaningful semantics across both modalities throughout the dataset. 
This approach ensures that the model captures essential temporal semantics relevant to patient outcomes, providing a more comprehensive understanding of the data.
To further enhance the quality of the learned temporal patterns, we introduce a Temporal Pattern Noise Contrastive Estimation (TP-NCE) loss for aligning pattern embeddings across modalities, along with auxiliary reconstruction losses to ensure that the patterns retain core information of the data.
Moreover, our framework incorporates a transformer~\cite{vaswani2017attention}-based fusion mechanism to effectively fuse the discovered temporal patterns with timestamp-level representations from both modalities, leading to more accurate predictions. 
We evaluate \modelname\ on two critical clinical prediction tasks: 48-hour in-hospital mortality prediction and 24-hour phenotype classification, using the MIMIC-III database.
The results demonstrate the effectiveness of our approach, which significantly outperforms existing methods, and suggest a promising direction for improving multimodal EHR analysis for clinical prediction.
}

\section{Related Works}
\subsection{EHR Time-Series Data Analysis}
EHR data is critical for clinical tasks such as disease diagnosis, mortality prediction, and treatment planning~\citep{harutyunyan2019multitask,zhang2023improving}. However, its high dimensionality and irregular nature pose challenges for traditional predictive models~\citep{rani2012recent,lee2017big}. Deep learning models, such as RNNs and LSTMs, are often used to capture temporal dependencies in EHR data~\citep{hayat2022medfuse,deldari2023latent}, but they struggle with irregular time intervals due to their reliance on fixed-length sequences~\citep{xie2021}.
To address this, some methods update patient representations at each time step using graph neural networks~\citep{zhang2021graph}, while others employ time-aware embeddings to incorporate temporal information~\citep{qian2023knowledge,zhang2023improving}. 
Despite these advancements, existing approaches still struggle to model high-level temporal patterns essential for accurate clinical outcome prediction.
%

\subsection{Prototype-based Pattern Learning}
Prototype-based learning identifies representative instances (prototypes) and optimizes their distance from input data in latent space for tasks like classification~\citep{li2023prototypes,ye2024ptarl}. This approach has been widely applied in tasks such as anomaly detection, unsupervised learning, and few-shot learning~\citep{tanwisuth2021prototype,li2023prototype}. Recently, it has been extended to time-series data~\citep{ghosal2021multi, li2023prototypes, yu2024imputation}, demonstrating its potential for detecting complex temporal patterns.
Additionally, prototype-based learning offers interpretable predictions, which is essential for healthcare applications~\citep{zhang2024prototypical}. However, learning efficient cross-modal temporal prototypes for multimodal EHR data remains an unexplored problem, as irregular time series and multi-scale patterns present significant challenges for existing methods.

\subsection{Multi-modal Learning in Healthcare}
In healthcare, patient data is typically collected in various forms—such as vital signs, laboratory results, medications, medical images, and clinical notes—to provide a comprehensive view of a patient’s health. Integrating these diverse modalities significantly enhances the performance of clinical tasks~\citep{hayat2022medfuse, zhang2023improving, yao2024drfuse}. However, fusing multimodal data remains challenging due to the heterogeneity and complexity of the sources.
Earlier research on multimodal learning~\citep{trong2020late, ding2022multimodal, hayat2022medfuse} often rely on late fusion strategies, where unimodal representations are combined via concatenation or Kronecker products. While straightforward, these approaches often fail to capture complex inter-modal interactions, leading to suboptimal representations.
Recent works have introduced transformer-based models that focus on cross-modal token interactions~\citep{zhang2023improving, theodorou2024framm,yao2024drfuse}. While these models are effective at capturing inter-modal relationships, they often struggle to extract high-level temporal semantics from multimodal data, limiting the ability to achieve a comprehensive understanding of a patient's health conditions.  
\section{Methodology}

\begin{figure*}[t!]
    \centering
    \includegraphics[width=0.95\textwidth]{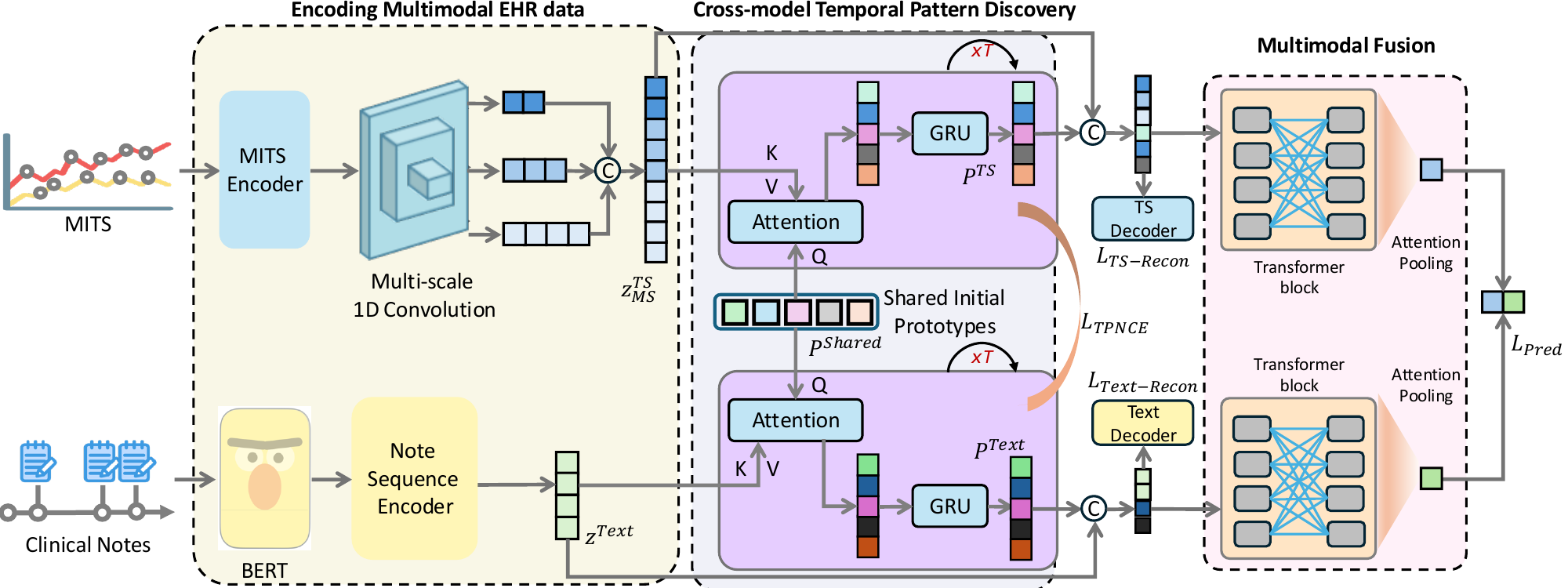}
    \caption{
    \textbf{\modelname\ overview}: the input Multivariate Irregular Time Series (MITS) and clinical note sequences are first encoded into regular embeddings. We then introduce the Cross-Modal Temporal Pattern Discovery (CTPD) module to extract meaningful temporal semantics. The extracted temporal patterns, along with the timestamp-level embeddings from both modalities, are fused to generate the final predictions.
    }
    \label{fig:framework}
    \vspace{-0.4cm}
\end{figure*}


\subsection{Problem Formulation}
\label{sec:problem_formulation}
In practice, multimodal EHR datasets contain multiple data types, specifically Multivariate Irregular Time Series (MITS) and free-text clinical notes.
We represent the multimodal EHR data for the $i$-th admission as $\{ (\mathbf{x}_{(i)}, \mathbf{t}^{\mathrm{TS}}_{(i)}), (n_{(i)}, \mathbf{t}^{\mathrm{Text}}_{(i)}), \mathbf{y}_{(i)} \}_{i=1}^{N}$.
Here $\mathbf{x}_{(i)}$ represents the multivariate time series observations, with $\mathbf{t}_{(i)}^{\mathrm{TS}}$ indicating their corresponding time points.
The sequence of clinical notes is represented by $n_{(i)}$, and $\mathbf{t}_{(i)}^{\mathrm{Text}}$ denotes the time points of these notes.
The variable $\mathbf{y}_i$ denotes the clinical outcomes to predict.
For simplicity, we omit the admission index $i$ in subsequent sections.
The MITS $\mathbf{x}$ comprises $d_m$ variables, where each variable $j = 1, ..., d_m$ has $l_{(j)}^{TS}$ observations, with the rest missing.
Similarly, each clinical note sequence $n$ includes $l^{Text}$ notes.
Early-stage medical prediction tasks aim to forecast an outcome $y$ for the admission $i$ using their multimodal EHR data $\{(\mathbf{x}, \mathbf{t}^{\mathrm{TS}}), (n, \mathbf{t}^{\mathrm{Text}})\}$, specifically before a certain time point (e.g., 48 hours) after admission.
Here, both $\mathbf{t}^{\mathrm{TS}}$ and $\mathbf{t}^{\mathrm{Text}}$ are constrained within a fixed temporal window following admission (e.g., 48 or 24 hours). Apart from this shared temporal boundary, the two sets of timestamps are recorded independently and are not inherently correlated.

\subsection{Encoding MITS and Clinical Notes}
\label{sec:encoding}
Here, we introduce our time series encoder $E_{\mathrm{TS}}$ and text encoder $E_{\mathrm{Text}}$, which separately encode MITS $(\mathbf{x}, \mathbf{t}^{\mathrm{TS}})$ and clinical notes $(n, \mathbf{t}^{\mathrm{Text}})$ into their respective embeddings $\mathbf{z}^{\mathrm{TS}}$ and $\mathbf{z}^{\mathrm{Text}} \in \mathbb{R}^{T\times D}$. Here $T$ denotes the number of regular time points, and $D$ denotes the embedding dimension.

For MITS, we utilize a gating mechanism that dynamically integrates both irregular time series embeddings $\mathbf{e}_{\mathrm{imp}}^{\mathrm{TS}}$ and imputed regular time series embeddings $\mathbf{e}_{\mathrm{mTAND}}^{\mathrm{TS}}$, following the approach in~\cite{zhang2023improving}.
Formally, the MITS embedding $\mathbf{z}^{\mathrm{TS}}$ is computed as:
\begin{equation}
\label{eq:gating}
\begin{aligned}
\mathbf{z}^{\mathrm{TS}} = \mathbf{g} \odot \mathbf{e}_{\mathrm{imp}}^{\mathrm{TS}} + (1 - \mathbf{g}) \odot \mathbf{e}_{\mathrm{mTAND}}^{\mathrm{TS}}
\end{aligned}
\tag{1}
\end{equation}
where $ \mathbf{g} = f(\mathbf{e}_{\mathrm{imp}}^{\mathrm{TS}} \oplus \mathbf{e}_{\mathrm{mTAND}}^{\mathrm{TS}}) \in \mathbb{R}^{T \times D}$, $f(\cdot)$ is a gating function implemented via an MLP, $\oplus$ denotes the concatenation, and $\odot$ denotes point-wise multiplication.

The regular time series $\mathbf{e}_{\mathrm{imp}}^{\mathrm{TS}}$ embedding is derived by applying a 1D convolution layer to the imputed time series.
At each reference time point $\alpha = 1, ..., T$, the imputed values are sourced from the nearest preceding values or replaced with a standard normal value if no prior data is available.
%
Concurrently, mTAND (multi-time attention)~\cite{shukla2021multi} generates an alternative set of time series representations $\mathbf{e}_{\mathrm{mTAND}}^{\mathrm{TS}}$ with the same reference time points $\mathbf{r}$ with irregular time representations.
Specifically, we leverage $V$ different Time2Vec~\cite{kazemi2019time2vec} functions $\{\theta_v(\cdot)\}_{v=1}^V$ to produce interpolation embeddings
at each time point $\alpha$, which are then concatenated and linearly projected to form $\mathbf{e}_{\mathrm{mTAND}}^{\mathrm{TS}}(\alpha) \in \mathbb{R}^{D}$. 

\revise{
For clinical notes, 
%
each of them is first encoded using Bert-tiny (a text encoder), where we extract the \texttt{[CLS]} token to represent this as a sequence of note-level embeddings $\mathbf{e}^{\mathrm{Text}} \in \mathbb{R}^{T \times D}$.
To tackle irregularity, we sort $\mathbf{e}^{\mathrm{Text}}$ according to their timestamps $\mathbf{T}^{\mathrm{Text}}$ and treat the pair $(\mathbf{e}^{\mathrm{Text}}, \mathbf{T}^{\mathrm{Text}})$ as a multimodal time series. Following the approach for time-series data, we leverage $V$ different Time2Vec features $\{\theta_v\}_{v=1}^V$ to produce interpolation embeddings at each time point $\alpha $:
\begin{equation}
\label{eq: text rep}
\begin{aligned}
\mathbf{z}^{\mathrm{Text}} = \text{mTAND}(\alpha, \mathbf{T}^{\mathrm{Text}}, \mathbf{e}^{\mathrm{Text}})
\end{aligned}
\tag{2}
\end{equation}
This yields temporally aligned text embeddings at each time point $ \alpha $, enabling effective interaction with time series data.
Note that our choice of BERT-Tiny is based on prior work~\cite{park2022graph} in multimodal EHR analysis, where its lightweight architecture has proven effective.
Nonetheless, our framework can be seamlessly integrated with other text encoders.
}

\subsection{Discover Cross-modal Temporal Patterns from Multimodal EHR}
\label{sec:dtp}


High-level temporal patterns in multimodal EHR data often encode rich medical condition-related semantics that are crucial for predicting clinical outcomes. However, previous works primarily focus on timestamp-level embeddings, frequently overlooking these important temporal patterns~\citep{bahadori2019temporal,xiao2023gaformer,sun2024time}. Drawing inspiration from object-centric learning in the computer vision domain~\citep{locatello2020object,li2021scouter}, we propose a novel temporal pattern discovery module to capture complex patterns within longitudinal data.

Considering the hierarchical nature~\cite{yue2022ts2vec,cai2024msgnet} of time series data, the critical temporal patterns for EHR may manifest across multiple time scales. Consequently, our approach performs temporal pattern discovery on multi-scale time series embeddings. 

\para{Extracting Cross-modal Temporal Patterns.}
Owing to the correspondence within multi-modal data, our cross-modal temporal pattern discovery module focuses on extracting corresponding temporal patterns across both modalities for a better understanding of multimodal EHR.
%
Starting with the time series embeddings $\mathbf{z}^{\mathrm{TS}}$ in Eq.~\ref{eq:gating}, we generate multi-scale embeddings (each scale can be divided by T)
$\{\mathbf{z}^{\mathrm{TS}}_{(1)}, \mathbf{z}^{\mathrm{TS}}_{(2)}, \mathbf{z}^{\mathrm{TS}}_{(3)}\}$ using three convolutional blocks followed by mean pooling with a stride of 2 along the time dimension.
Each resulting embedding corresponds to a different temporal scale, where the sequence lengths are fractions of the original length $T$ (e.g., $T$, $T/2$, and $T/4$).
The concatenated multi-scale embedding $\mathbf{z}^{\mathrm{TS}}_{\mathrm{MS}} \in \mathbb{R}^{T' \times D}$ serves as the diverse temporal representation, where $T' = T + T/2 + T/4 = 1.75T$.
Since $T$ is set to either 48 or 24 in our experiments, $T'$ remains an integer.
We then enhance these embeddings by applying position encoding:
$\mathbf{\hat{z}}^{\mathrm{TS}}_{\mathrm{MS}} = \mathbf{z}^{\mathrm{TS}}_{\mathrm{MS}} + \mathrm{PE}(\mathbf{z}^{\mathrm{TS}}_{\mathrm{MS}})$, $\mathbf{\hat{z}}^{\mathrm{TS}}_{\mathrm{MS}} \in \mathbb{R}^{T' \times D}$,
where $\mathrm{PE}(\cdot)$ denotes the position embeddings in~\citep{vaswani2017attention}.
Furthermore, to capture potential temporal patterns, we define a group of $K$ learnable vectors as temporal prototypes, $\mathbf{P}^{\mathrm{Shared}} \in \mathbb{R}^{K \times D}$, initially sampled from a normal distribution $ \mathcal{N}(\mu, \operatorname{diag}(\sigma)) \in \mathbb{R}^{K \times D}$ and refined during training.
%
The shared prototype embeddings are designed to capture semantic-corresponding temporal patterns across modalities, respectively, with $\mu$ and $\sigma$ randomly initialized and subsequently optimized.

%
To extract temporal patterns, we first calculate the assignment weights $\mathbf{W} $  between prototype embeddings and modality embeddings using a dot-product attention mechanism:
\begin{equation}
\begin{aligned}
\mathbf{W}^{\mathrm{TS}} &= \mathrm{Attention}(\mathbf{P}^{\mathrm{Shared}}, \mathbf{\hat{z}}^{\mathrm{TS}}_{\mathrm{MS}}), \\
\mathbf{W}^{\mathrm{Text}} &= \mathrm{Attention}(\mathbf{P}^{\mathrm{Shared}}, \mathbf{z}^{\mathrm{Text}})
\end{aligned}
\tag{3}
\end{equation}
The $\mathrm{Attention}$ mechanism is defined as:
\begin{equation}
\begin{aligned}
\mathrm{Attention}(\mathbf{q}, \mathbf{k})_{i, j}&=\frac{e^{M_{i, j}}}{\sum_l e^{M_{i, l}}},\notag\\ 
\end{aligned}
\tag{4}
\end{equation}
where $\textbf{M}=\frac{1}{\sqrt{D}}g_q(\mathbf{q}) \cdot g_k(\mathbf{k})^T $ and $g_q(\cdot)$ and $g_k(\cdot)$ are two learnable matrices.
Next, we aggregate the input values to their assigned prototypes using a weighted mean to obtain updated embeddings $\mathbf{z}^{\mathrm{TS}}_{\mathrm{updated}}$ and $\mathbf{z}^{\mathrm{Text}}_{\mathrm{updated}}$:
\begin{equation}
\begin{aligned}
\mathbf{z}^{\mathrm{TS}}_{\mathrm{updated}} &= \mathbf{W}^{\mathrm{TS}} \cdot v(\mathbf{\hat{z}}^{\mathrm{TS}}_{\mathrm{MS}}) \in \mathbb{R}^{K \times D},  \\
\mathbf{z}^{\mathrm{Text}}_{\mathrm{updated}} &= \mathbf{W}^{\mathrm{Text}} \cdot v(\mathbf{\hat{z}}^{\mathrm{Text}}) \in \mathbb{R}^{K \times D}, 
\notag \\
\end{aligned}
\tag{5}
\end{equation}
where $v(\cdot)$ is a learnable matrix. 

Finally, the prototype embeddings $\mathbf{P}^{\mathrm{TS}}$ and $\mathbf{P}^{\mathrm{Text}}$ are refined using the corresponding updated embeddings via a learned recurrent function: 
\begin{equation}
\begin{aligned}
\mathbf{P}^{\mathrm{TS}} &= f(\mathrm{GRU} (\mathbf{P}^{\mathrm{TS}}_{(0)})) \in \mathbb{R}^{K\times D}
\notag \\
\mathbf{P}^{\mathrm{Text}} &= f(\mathrm{GRU} (\mathbf{P}^{\mathrm{Text}}_{(0)})) \in \mathbb{R}^{K\times D}
\end{aligned}
\tag{6}
\end{equation}
where $\mathbf{P}^{\mathrm{TS}}_{(0)}$ and $\mathbf{P}^{\mathrm{Text}}_{(0)}$ are the prototype embeddings from the previous step 
, $\mathrm{GRU}(\cdot)$ is Gated Recurrent Unit~\citep{cho2014learning}, and $f(\cdot)$ denotes MLP. The above process is repeated for 3 iterations per step.
Those refined embeddings denote the discovered temporal patterns for each modality.
\revise{
Note that our design of attention and GRU is inspired by Slot Attention~\cite{locatello2020object}, which has been shown to be effective in learning object-centric representations from images.}

\para{TP-NCE Contraint.}
To ensure consistent semantics across modalities, we introduce a Temporal Pattern Noise Contrastive Estimation (TP-NCE) loss, inspired by InfoNCE~\cite{oord2018representation}, to enforce the similarity of multimodal prototype embeddings for the same ICU stay while increasing the distance between prototype embeddings from different ICU stays. 
\revise{
For a minibatch of $B$ samples, the TP-NCE loss from MITS to notes is defined as:
\begin{equation}
\begin{aligned}
\mathcal{L}_{\mathrm{TPNCE}}^{\mathrm{TS} \rightarrow \mathrm{Text}} = -\sum_{i=1}^{B}\big(\mathrm{log}\frac{\mathrm{exp}(\mathrm{sim}(i, i) / \tau)}{\sum_{j=1}^{B}\mathrm{exp}(\mathrm{sim}(i, j) / \tau)} \big)
\end{aligned}
\tag{7}
\end{equation}
where $\tau$ is a temperature parameter, and $i,j$ ($1 \leq i, j \leq B$) denote sample indices within the minibatch.
The similarity function $\mathrm{sim}(i, j)$ measures the similarity between the $i$-th $\mathbf{P}^{\mathrm{TS}}$ and $j$-th $\mathbf{P}^{\mathrm{Text}}$, and is defined as (for convenience, we omit the indices $i$ and $j$ in the equation below):
\begin{equation}
\begin{aligned}
\mathrm{sim}(\cdot) = \sum_{k=1}^{K}(\beta_k <\mathbf{P}^{\mathrm{TS}}(k), \mathbf{P}^{\mathrm{Text}}(k)>)
\end{aligned}
\tag{8}
\end{equation}
where $\langle \cdot \rangle$ denotes cosine similarity, and $k$ is the prototype index.
The bidirectional TP-NCE loss is then given by:
$\mathcal{L}_{\mathrm{TPNCE}} = \frac{1}{2}(\mathcal{L}_{\mathrm{TPNCE}}^{\mathrm{TS} \rightarrow \mathrm{Text}} + \mathcal{L}_{\mathrm{TPNCE}}^{\mathrm{Text} \rightarrow \mathrm{TS}})$.
}
To account for varying prototype importance, an attention mechanism is used to generate weights $\mathbf{\beta}$ for the slots, based on global MITS and text embeddings:
$\mathbf{\beta} = \mathrm{MLP}(\mathrm{concat}[\mathbf{g}^{\mathrm{TS}}_{\mathrm{MS}}, \mathbf{g}^{\mathrm{Text}}])$,
where $\mathbf{g}^{\mathrm{TS}}_{\mathrm{MS}}$ and $\mathbf{g}^{\mathrm{Text}} \in \mathbb{R}^{D}$ are global embeddings obtained by averaging $\mathbf{\hat{z}}^{\mathrm{TS}}_{\mathrm{MS}}$ and $\mathbf{z}^{\mathrm{Text}}$ along the time dimension.

\para{Auxiliary Reconstruction.}
To ensure that the learned prototype representations capture core information from multimodal EHR data, we introduce two reconstruction objectives aimed at reconstructing imputed regular time series and text embeddings from the learned prototypes. Specifically, we implement a time series decoder to reconstruct the imputed regular time series from $\mathbf{P}^{\mathrm{TS}}$, and a text embedding decoder to reconstruct text embeddings from $\mathbf{P}^{\mathrm{Text}}$. Both decoders are based on a transformer decoder architecture~\cite{vaswani2017attention}, and two mean squared error (MSE) losses denoted by $\mathcal{L}_{\mathrm{TS-Recon}}$ and $\mathcal{L}_{\mathrm{Text-Recon}}$ are used as the objective function.
Here, we define the overall reconstruction loss $\mathcal{L}_{\mathrm{Recon}} = \frac{1}{2}(\mathcal{L}_{\mathrm{TS-Recon}} + \mathcal{L}_{\mathrm{Text-Recon}})$.

\subsection{Multimodal Fusion}
\label{sec:fusion}
%
%
Since information from both modalities is crucial for predicting medical conditions, we propose a multimodal fusion mechanism to integrate these inputs.
%
%
\revise{
First, we apply a 2-layer transformer encoder~\cite{vaswani2017attention} to capture interactions between timestamp-level and prototype embeddings across both modalities for each sample.
We continue to use $\mathbf{P}^{\mathrm{TS}}$ and $\mathbf{P}^{\mathrm{Text}}$ to represent the resulted prototype embeddings for time-series data and clinical notes, respectively.
Similarly, $\hat{\mathbf{z}}^{\mathrm{TS}}_{\mathrm{MS}}$ and $\mathbf{z}^{\mathrm{Text}}$ denote the corresponding timestamp-level embeddings.
Then, we aggregate $K$ prototype embeddings and $T$ timestamp-level embeddings of each modality using an attention-based pooling mechanism:
\begin{equation}
\begin{aligned}
\mathbf{F}^{\mathrm{TS}} &= \sum_{k=1}^{K} \mathbf{\gamma}_{k}^{\mathrm{TS}} \mathbf{P}^{\mathrm{TS}}(k) + \sum_{t=1}^{T} \mathbf{\phi}_{t}^{\mathrm{TS}}\hat{\mathbf{z}}^{\mathrm{TS}}_{\mathrm{MS}}(t) \\
\mathbf{F}^{\mathrm{Text}} &= \sum_{k=1}^{K} \mathbf{\gamma}_{k}^{\mathrm{Text}} \mathbf{P}^{\mathrm{Text}}(k) + \sum_{t=1}^{T} \mathbf{\phi}_{t}^{\mathrm{Text}}\mathbf{z}^{\mathrm{Text}}(t)
\end{aligned}
\tag{9}
\end{equation}
Here $k$ and $t$ refer to the indices of prototype embeddings and timestamp-level embeddings, respectively.
Here $\mathbf{\gamma}^{\mathrm{TS}}$, $\mathbf{\phi}^{\mathrm{TS}}$, $\mathbf{\gamma}^{\mathrm{Text}}$, $\mathbf{\phi}^{\mathrm{Text}}$ are learned attention weights by passing the corresponding embeddings through a shared MLP.
}
The resulting global embeddings from both modalities are concatenated along the feature dimension to form the final global representation.

\subsection{Overall Learning Objectives}
\label{sec:objective}
To optimize our framework, we employ four loss functions jointly.
%
The overall objective is a weighted sum of these loss functions:
\begin{equation}
\begin{aligned}
\mathcal{L} = &\mathcal{L}_{\mathrm{pred}} + \lambda_1 * \mathcal{L}_{\mathrm{TPNCE}} + \lambda_2 * \mathcal{L}_{\mathrm{Recon}}
\end{aligned}
\tag{10}
\end{equation}
where $\lambda_1$, $\lambda_2$ are hyperparameters that control the weights of respective losses.
Here $\mathcal{L}_{\mathrm{pred}}$, is a cross-entropy loss used for classification.

\section{Experiment}

\subsection{Experimental Setup}
\label{sec:experiment_setup}

\noindent\textbf{Dataset.}
\revise{
We assess our model's efficacy using MIMIC-III v1.4\footnote{\url{https://physionet.org/content/mimiciii/1.4/}}, a comprehensive open-source multimodal clinical database~\citep{johnson2016mimic}.
We focus our evaluation on two critical tasks, \textit{48-hour in-hospital mortality prediction} (48-IHM) and \textit{24-hour phenotype classification} (24-PHE), as established in prior research~\citep{zhang2023improving,hayat2022medfuse}. 
%
}
\revise{
We extracted raw, irregular time-series data (containing 17 clinical variables) from the MIMIC-III database and selected time series within the first 48-hour and 24-hour windows within each ICU stay for each respective task, as done in~\cite{zhang2023improving,hayat2022medfuse,harutyunyan2019multitask}. 
ICU stays shorter than 48 or 24 hours were excluded from our dataset, which explains the different sample size between these two tasks.
Unlike~\cite{harutyunyan2019multitask}, we did not apply imputation during preprocessing but instead retained the original irregular time-series structure, following~\cite{zhang2023improving}.
Following the dataset splitting by~\citep{harutyunyan2019multitask}, we ensure that the model evaluation is robust by partitioning the data into $70\%$ training, $10\%$ validation, and $20\%$ testing sets, based on unique subject IDs to prevent information leakage.
For multimodal analysis, we paired numerical time-series data with corresponding clinical notes from each patient’s ICU stay, consistent with~\cite{zhang2023improving}.
%
%
Note that the pipeline of processing clinical notes we used follows the practice in~\cite{khadanga2019using}.
The dataset statistics, including sample counts before and after multimodal paring, are presented in Table~\ref{tab:dataset_statistics}.
The additional details of experimental setup can be found in Appendix~\ref{sec:dataset_details}.
}

\begin{table}[t!]
  \centering
  \caption{
  \revise{
    Number of samples for EHR and fully paired EHR-clinical notes across the training, validation, and test sets.
  }
  }
  \resizebox{0.75\linewidth}{!}{
  \begin{tabular}{c | c c c}
  \toprule
      & Training & Validation & Test \\
  \midrule
  \multicolumn{4}{c}{\textit{Number of EHR.}} \\
  \midrule
  48-IHM & 16,093 & 1,810 & 3,236 \\
  24-PHE & 26,891 & 2,955 & 5,282 \\
  \midrule
  \multicolumn{4}{c}{\textit{Number of Paired EHR and Clinical Notes.}} \\
  \midrule
  48-IHM & 15,425 & 1,727 & 3,107 \\
  24-PHE & 25,435 & 2,807 & 5,013 \\
  \bottomrule
  \end{tabular}
  }
  \label{tab:dataset_statistics}
  \vspace{-0.5cm}
\end{table}

\colorlet{shadecolor}{gray!10}
\begin{table*}[ht!]
    \centering
    \caption{
    Comparison of our method with baselines on 48-IHM and 24-PHE tasks using the MIMIC-III dataset.
    We report average performance on three random seeds, with standard deviation as the subscript.
    The \textbf{Best} and \underline{2nd best} methods under each setup are bold and underlined.
    %
    %
    }
    \resizebox{0.9\linewidth}{!}{
    \begin{tabular}{c | c c c |c c c}
        \toprule
        \multirow{2}{*}{Model} & \multicolumn{3}{c|}{48-IHM} & \multicolumn{3}{c}{24-PHE} \\
        & AUROC ($\uparrow$) & AUPR ($\uparrow$)  & F1 ($\uparrow$)  & AUROC  ($\uparrow$) & AUPR ($\uparrow$) & F1 ($\uparrow$)  \\
        \midrule
        \multicolumn{7}{c}{
        \textit{Methods on MITS.}
        } \\
        \midrule
        CNN~\citep{lecun1998gradient} & 85.80$_\textit{0.32}$ & 49.73$_\textit{0.65}$ & 46.37$_\textit{3.01}$ &
        75.36$_\textit{0.18}$ & 38.10$_\textit{0.26}$ &
        40.80$_\textit{0.37}$\\
        RNN~\citep{elman1990finding} & 84.75$_\textit{0.74}$ & 46.57$_\textit{1.18}$ & 45.60$_\textit{2.06}$ & 
        73.78$_\textit{0.10}$ & 36.76$_\textit{0.27}$ & 
        33.99$_\textit{0.48}$\\
        LSTM~\cite{hochreiter1997long} & 85.22$_\textit{0.67}$ &
        46.93$_\textit{1.28}$ & 45.72$_\textit{1.41}$ &
        74.46$_\textit{0.23}$ & 36.80$_\textit{0.28}$ & 
        39.45$_\textit{0.49}$\\
        Transformer~\citep{vaswani2017attention} & 83.45$_\textit{0.97}$ & 43.03$_\textit{1.65}$ & 
        39.31$_\textit{3.83}$ & 74.98$_\textit{0.14}$ & 
        39.37$_\textit{0.26}$ & 36.13$_\textit{1.42}$\\
        IP-Net~\citep{shukla2019interpolation} & 81.76$_\textit{0.38}$ & 39.50$_\textit{0.83}$ & 43.89$_\textit{1.07}$ & 
        73.98$_\textit{0.13}$ & 35.31$_\textit{0.29}$ & 
        39.38$_\textit{0.15}$\\
        GRU-D~\citep{che2018recurrent} & 49.21$_\textit{5.26}$ & 12.85$_\textit{2.24}$ & 19.63$_\textit{0.01}$ & 52.11$_\textit{0.42}$ & 17.99$_\textit{0.55}$ & 26.17$_\textit{0.23}$\\
        DGM-O~\citep{wu2021dynamic} & 
        71.99$_\textit{7.30}$ & 28.67$_\textit{7.34}$ & 
        31.72$_\textit{11.02}$ & 60.70$_\textit{0.82}$ & 
        22.56$_\textit{0.77}$ & 28.87$_\textit{0.64}$\\
        mTAND~\citep{shukla2021multi} & 85.27$_\textit{0.20}$ & 
        49.82$_\textit{0.97}$ & 48.02$_\textit{1.93}$ & 
        72.79$_\textit{0.09}$ & 35.95$_\textit{0.14}$ & 
        32.42$_\textit{0.72}$\\
        SeFT~\citep{horn2020set} & 65.00$_\textit{0.84}$ & 
        22.93$_\textit{1.27}$ & 19.70$_\textit{15.59}$ & 
        60.50$_\textit{0.09}$ & 23.57$_\textit{0.07}$ & 
        21.26$_\textit{0.21}$\\
        UTDE~\citep{zhang2023improving} & 86.14$_\textit{0.45}$ & 50.60$_\textit{0.28}$ & 49.29$_\textit{0.62}$ & 73.62$_\textit{0.57}$ & 36.80$_\textit{0.87}$ & 40.58$_\textit{0.64}$\\
        \midrule
        \multicolumn{7}{c}{
        \textit{Methods on Clinical Notes.}
        } \\
        \midrule
        Flat~\cite{deznabi2021predicting} & 85.71$_\textit{0.49}$ & 50.96$_\textit{0.19}$ & 45.80$_\textit{6.61}$ & 
        81.77$_\textit{0.09}$ & 53.79$_\textit{0.21}$ & 
        \underline{52.13$_\textit{0.79}$} \\
        HierTrans~\cite{pappagari2019hierarchical} & 84.32$_\textit{0.34}$ & 
        47.92$_\textit{0.57}$ & 42.01$_\textit{0.58}$ & 80.79$_\textit{0.06}$ & 
        51.97$_\textit{0.08}$ & 50.85$_\textit{0.71}$\\
        T-LSTM~\cite{baytas2017patient} &
        85.70$_\textit{0.21}$ & 45.29$_\textit{0.86}$ &
        42.84$_\textit{5.53}$ & 81.15$_\textit{0.04}$ &
        50.23$_\textit{0.07}$ & 49.77$_\textit{0.24}$\\
        FT-LSTM~\cite{zhang2020time} & 
        84.28$_\textit{0.95}$ & 43.93$_\textit{1.03}$ &
        36.87$_\textit{6.54}$ & 81.66$_\textit{0.12}$ &
        51.71$_\textit{0.24}$ & 50.21$_\textit{0.85}$ \\
        GRU-D~\citep{che2018recurrent} & 72.58$_\textit{0.42}$ & 26.38$_\textit{1.06}$ & 31.64$_\textit{0.79}$ & 49.52$_\textit{0.65}$ & 16.90$_\textit{0.12}$ & 27.80$_\textit{0.08}$\\
        mTAND~\citep{shukla2021multi} & 85.40$_\textit{0.60}$ & 49.68$_\textit{1.26}$ & 35.76$_\textit{11.57}$ & \underline{82.14$_\textit{0.07}$} & \underline{54.57$_\textit{0.15}$} & 52.01$_\textit{0.93}$\\
        \midrule
        \multicolumn{7}{c}{
        \textit{Methods on Multimodal EHR.}
        } \\
        \midrule
        MMTM~\citep{joze2020mmtm} & \underline{87.88$_\textit{0.07}$} & \underline{53.58$_\textit{0.24}$} & 51.54$_\textit{1.58}$ & 81.46$_\textit{0.25}$ & 51.88$_\textit{0.12}$ & 51.59$_\textit{0.19}$\\
        DAFT~\citep{polsterl2021combining} & 87.53$_\textit{0.22}$ & 52.40$_\textit{0.21}$ & \underline{51.95$_\textit{0.64}$} & 81.18$_\textit{0.08}$ & 50.91$_\textit{0.31}$ & 50.72$_\textit{0.39}$\\
        MedFuse~\citep{hayat2022medfuse} & 86.02$_\textit{0.29}$ & 51.00$_\textit{0.22}$ & 49.29$_\textit{0.75}$ & 78.88$_\textit{0.14}$ & 45.99$_\textit{0.21}$ & 47.47$_\textit{0.23}$\\
        DrFuse~\citep{yao2024drfuse} & 85.97$_\textit{1.02}$ & 49.94$_\textit{1.91}$ & 49.75$_\textit{1.52}$ & 80.88$_\textit{0.18}$ & 49.62$_\textit{0.40}$ & 50.18$_\textit{0.31}$ \\
        \rowcolor{Gray}
        \modelname \ (Ours) & \textbf{88.15$_\textit{0.28}$} & \textbf{53.86$_\textit{0.65}$} & \textbf{53.85$_\textit{0.16}$} & \textbf{83.34$_\textit{0.05}$} & \textbf{56.39$_\textit{0.17}$} & \textbf{53.83$_\textit{0.43}$} \\
        \bottomrule
        \end{tabular}
    }
    \label{tab:comparison_with_sota_mimic_iii}
    \vspace{-0.3cm}
\end{table*}

\begin{table*}[t!]
  \centering
  \caption{
    Statistics analysis of \modelname\ on MIMIC-III dataset.
    p-values are computed from paired t-tests. 
  }
  \resizebox{0.7\linewidth}{!}{
  \begin{tabular}{c | c c c | c c c}
      \toprule
      \multirow{2}{*}{Model} & \multicolumn{3}{c|}{48-IHM} & \multicolumn{3}{c}{24-PHE} \\
      & AUROC ($\uparrow$) & AUPR ($\uparrow$)  & F1 ($\uparrow$)  & AUROC  ($\uparrow$) & AUPR ($\uparrow$) & F1 ($\uparrow$)  \\
      \midrule
      SOTA & 87.88$_\textit{0.07}$ & 53.58$_\textit{0.24}$ & 51.95$_\textit{0.64}$ & 82.14$_\textit{0.07}$ & 54.57$_\textit{0.15}$ & 52.13$_\textit{0.79}$ \\
      CTPD (Ours) & 88.15$_\textit{0.28}$ & 53.86$_\textit{0.65}$ &    53.85$_\textit{0.16}$ & 83.34$_\textit{0.05}$ & 56.39$_\textit{0.17}$ & 53.83$_\textit{0.43}$ \\
      \rowcolor{Gray}
        Gains & \highlightgain{+0.27} & \highlightgain{+0.28} & \highlightgain{+1.9} & \highlightgain{+1.2} & \highlightgain{+1.82} & \highlightgain{+1.7} \\
      p values & 0.016 & 0.233 & 7.74$\mathrm{e}{-6}$ & 7.89$\mathrm{e}{-12}$ &  1.10$\mathrm{e}{-10}$ & 2.08$\mathrm{e}{-4}$ \\
      \bottomrule
      \end{tabular}
  }
  \label{tab:statistic_analysis}
  \vspace{-0.3cm}
\end{table*}
\para{Evaluation Metrics.}
The 48-hour In-Hospital Mortality (48-IHM) prediction is a binary classification with a marked label imbalance, indicated by a death-to-discharge ratio of approximately 1:6.
Following previous work~\citep{harutyunyan2019multitask,zhang2023improving}, we use AUROC, AUPR, and F1 score, for a comprehensive evaluation.
The $24$-hour Phenotype Classification (24-PHE) involves predicting the presence of 25 different medical conditions during an ICU stay, making it a multi-label classification task.
For this task, we employ the AUROC, AUPR, and F1 score (Macro) for a thorough assessment of model efficacy.
The F1 score threshold is determined by selecting the value that maximizes the F1 score on the validation set.

\para{Implementation Details.}
%
%
We train the model with batch size of 128, learning rate of 4e-5, and Adam~\citep{kingma2014adam} optimizer.
We use a cosine annealing learning rate scheduler with a 0.2 warm-up proportion.
To prevent overfitting, we implement early stopping when there is no increase in the AUROC on the validation set for 48-IHM or 24-PHENO over $5$ consecutive epochs.
All experiments are conducted on 1 RTX-3090 GPU card using about 1 hour per run.
We clip the norm of gradient values with 0.5 for stable training.
By default, we use Bert-tiny~\cite{DBLP:journals/corr/abs-1908-08962,bhargava2021generalization} as our text encoder.

\para{Compared Methods.}
To ensure a comprehensive comparison, we compare our \modelname\ with three types baselines: MITS-only approaches, note-only approaches and multimodal approaches.
For MITS-only setting, we compare $\text{\modelname}$ with 4 baselines for imputed regular time series: RNN~\citep{elman1990finding}, LSTM~\citep{hochreiter1997long}, CNN~\citep{lecun1998gradient} and Transformer~\cite{vaswani2017attention}, and 5 baselines for irregular time series, including IP-Net~\citep{shukla2019interpolation}, GRU-D~\citep{che2018recurrent}, DGM-O~\citep{wu2021dynamic}, mTAND~\citep{shukla2021multi}, SeFT~\citep{horn2020set}, and UTDE~\citep{zhang2023improving}. The imputation approach follows the MIMIC-III benchmark~\citep{harutyunyan2019multitask}.
For the note-only setting, we compare our model with 6 baselines: Flat~\cite{deznabi2021predicting}, HierTrans~\cite{pappagari2019hierarchical}, T-LSTM~\cite{baytas2017patient}, FT-LSTM~\cite{zhang2020time}, GRU-D~\citep{che2018recurrent}, and mTAND~\citep{shukla2021multi}.
In the multimodal setting, we compare our model with 4 baselines: MMTM~\cite{joze2020mmtm}, DAFT~\cite{polsterl2021combining}, MedFuse~\cite{hayat2022medfuse}, and DrFuse~\cite{yao2024drfuse}.
To ensure a fair comparison, we implement Bert-tiny~\cite{bhargava2021generalization,DBLP:journals/corr/abs-1908-08962} as the text encoder across all baselines.
Details of baselines can be found in the Appendix \ref{appendix:more_baseline}.

\subsection{Comparison with SOTA Baselines}

\para{Results on MIMIC-III.}
Table~\ref{tab:comparison_with_sota_mimic_iii} presents a comparison of our proposed \modelname\ against 3 types of baselines: MITS-based methods, clinical notes-based methods, and multimodal EHR-based methods. 
Our \modelname, which incorporates cross-modal temporal pattern embeddings, consistently achieves the best performance across all 6 metrics.
Specifically, \modelname\ shows a $1.89\%$ improvement in F1 score on the 48-IHM task, and a $1.2\%$ improvement in AUROC and $1.92\%$ in AUPR on the more challenging 24-PHE task, compared to the second-best results.
\revise{
Additionally, we have conducted statistical analysis in Table~\ref{tab:statistic_analysis} to assess statistical significance.
\modelname\ demonstarted significant gains over SOTAs (p value $< 0.05$) in 5 out of 6 settings, indicating its effectiveness in analyzing multimodal EHR.
}

\para{Evaluation on More Tasks and Datasets.}
\revise{
To further evaluate the generalizability of our CTPD framework, we have extended it to another important admission-level task: 30-day readmission prediction on MIMIC-III.
Additionally, we also conducted experiments on the additional MIMIC-IV dataset to assess our framework's adaptability to different data sources.
These results can be found in Appendix~\ref{sec:readm} and Appendix~\ref{sec:mimic_iv}. 
}

\para{Discussion on Missing Modalities and Noisy Data Scenarios.}
\revise{
Currently, CTPD framework is built on paired time-series and clinical notes.
In practice, the dataset might have missing modalities, such as partial clinical notes or time-series data are missed. 
%
%
We acknowledge this is an interesting research direction.
However, the study of missing modalities falls outside the scope of our work and will be explored in future research.
Additionally, our evaluation is conducted on MIMIC-III~\cite{johnson2016mimic}, a large-scale, de-identified real-world dataset containing patient records from critical care units at Beth Israel Deaconess Medical Center between 2001 and 2012. Our approach is designed to be adaptable and can be seamlessly applied to other real-world databases.
}
%
%



\begin{table}[t!]
    \centering
    \caption{
    Ablation results show the impact of removing different types of input embeddings. 
    }
    \resizebox{\linewidth}{!}{
    \begin{tabular}{c | c c c | c c c c}
        \toprule
        & \multicolumn{3}{c|}{48-IHM} & \multicolumn{3}{c}{24-PHE} \\
        \midrule
         & AUROC & AUPR & F1 & AUROC & AUPR & F1 \\
        \midrule
        \rowcolor{Gray}
        Ours & \textbf{88.15$_\textit{0.28}$} & \textbf{53.86$_\textit{0.65}$} & \textbf{53.85$_\textit{0.16}$} & \textbf{83.34$_\textit{0.05}$} & \textbf{56.39$_\textit{0.17}$} & \textbf{53.83$_\textit{0.43}$} \\
        :w/o prototype & 86.89$_\textit{0.97}$ & 53.67$_\textit{0.65}$ & 48.47$_\textit{6.13}$ & 82.24$_\textit{0.07}$ & 54.06$_\textit{0.07}$ & 52.88$_\textit{0.11}$ \\
        :w/o timestamp embeddings & 87.18$_\textit{0.94}$ & 54.32$_\textit{1.66}$ & 45.85$_\textit{4.91}$ & 82.41$_\textit{0.15}$ & 54.30$_\textit{0.21}$ & 53.34$_\textit{0.41}$ \\
        :w/o multi-scale embedding & 87.59$_\textit{0.49}$ &53.38$_\textit{1.79}$ & 49.74$_\textit{3.50}$ & 83.11$_\textit{0.09}$ & 55.95$_\textit{0.05}$ & 53.81$_\textit{0.46}$ \\
        \bottomrule
    \end{tabular}
    \label{tab:ablation_slots}
    }
\end{table}

\begin{table}[t!]
    \centering
    \caption{
    Ablation study of loss functions $\mathcal{L}_{\mathrm{TPNCE}}$ and $\mathcal{L}_{\mathrm{Recon}}$. The \textbf{Best} results are highlighted in bold.
    }
    \vspace{-0.1cm}
    \resizebox{\linewidth}{!}{
    \begin{tabular}{c c | c c c | c c c}
        \toprule
        & & \multicolumn{3}{c|}{48-IHM} & \multicolumn{3}{c}{24-PHE} \\
        \midrule
        Cont & Recon & AUROC & AUPR & F1 & AUROC & AUPR & F1 \\
        \midrule  
        & & 87.49$_\textit{0.47}$ & 53.31$_\textit{1.46}$ & 43.59$_\textit{4.63}$ & 82.49$_\textit{0.10}$ & 55.25$_\textit{0.30}$ & 53.71$_\textit{0.43}$ \\
        \checkmark & & 87.15$_\textit{0.38}$ & 53.45$_\textit{1.09}$ & 44.76$_\textit{4.48}$ & 82.94$_\textit{0.05}$ & 55.62$_\textit{0.22}$ & 53.99$_\textit{0.19}$ \\
        & \checkmark & 86.93$_\textit{0.69}$ & 52.70$_\textit{1.97}$ & 41.52$_\textit{0.90}$ & 82.86$_\textit{0.03}$ & 55.43$_\textit{0.05}$ & \textbf{54.08$_\textit{0.29}$} \\
        \rowcolor{Gray}
        \checkmark & \checkmark & \textbf{88.15$_\textit{0.28}$} & \textbf{53.86$_\textit{0.65}$} & \textbf{53.85$_\textit{0.16}$} & \textbf{83.34$_\textit{0.05}$} & \textbf{56.39$_\textit{0.17}$} & 53.83$_\textit{0.43}$ \\
        \bottomrule
    \end{tabular}
    }
    \label{tab:ablation_loss}
    \vspace{-0.4cm}
\end{table}

\begin{table*}[ht!]
    \centering
    \caption{Number of parameters of different modules in \modelname.}
    \vspace{-0.2cm}
    \label{tab:number_of_parameters}
    \resizebox{0.9\textwidth}{!}{
    \begin{tabular}{c c c c c c c c c c}
        \toprule
        \textbf{Module} & Multi-scale Convolution & Shared Prototype & Dual Slot Attention & MITS Decoder & Text Decoder & Multimodal Fusion & Text Encoder & Others & \textbf{Total} \\
        \midrule
        \textbf{\#. Trainable Parameters} & 312.2K & 256 & 364K & 1.3M & 1.3M & 1.23M & 4.4M & 0.294M & \textbf{9.4M} \\
        \bottomrule
    \end{tabular}
    }
    \vspace{-0.5cm}
\end{table*}

\subsection{Model Analysis}

\para{Ablation Results on Different Components.} 
%
%
We conduct ablation studies by removing the prototype embeddings, timestamp-level embeddings, and multi-scale feature extractor respectively, and analyze their impacts on two clinical prediction tasks, as shown in Table~\ref{tab:ablation_slots}. Notably, prototype embeddings play the most significant role among the three components, with their removal resulting in a $1.96\%$ AUROC decrease in 48-IHM and a $1.1\%$ decrease in 24-PHE. 
The results also show that all three embeddings are important for capturing effective information for prediction.

\para{Ablation Results on Learning Objectives.}
Table~\ref{tab:ablation_loss} presents the ablation results of the learning objectives. Combining both $\mathcal{L}_{\mathrm{TPNCE}}$ and $\mathcal{L}_{\mathrm{Recon}}$ leads to the best performance across 5 out of 6 settings. 
Our model remains robust under different loss function configurations, with only a $0.66\%$ AUROC drop in the model's performance in 48-IHM and a $0.85\%$ drop in 24-PHE.
Note that the first row of Table~\ref{tab:ablation_loss}, i.e., \modelname\ without auxiliaries losses performs slightly worse than several SOTAs (Table~\ref{tab:comparison_with_sota_mimic_iii}) in 48-IHM task, while significantly outperforms compared approaches in 24-PHE task.
We think The performance gap is mainly due to task complexity. The 48-IHM task is a simple binary classification problem with limited need for cross-modal reasoning, making our model more prone to overfitting due to its added components. In contrast, the 24-PHE task is a more complex multi-label classification problem that benefits from stronger temporal and cross-modal understanding, leading to larger performance gains over SOTAs.

\para{Complexity of the Proposed Method}
\label{sec:complexity}
To provide a more detailed analysis of the computational complexity of our proposed framework, we report the number of trainable parameters for each module in the Table~\ref{tab:number_of_parameters}. Our overall model only has 9.2 M parametes, which is lightweight for multimodal EHR analysis.
In addition, we evaluated the inference speed of our \modelname\ and found that it requires only \textbf{5.42 ms} per time series-text pair on average. This indicates the framework’s potential for real-time deployment in clinical settings.
%

\begin{figure}[t!]
    \centering
    \caption{
        Ablation study on the number of prototypes. 
    }
    \vspace{-0.2cm}
    \includegraphics[width=0.85\linewidth]{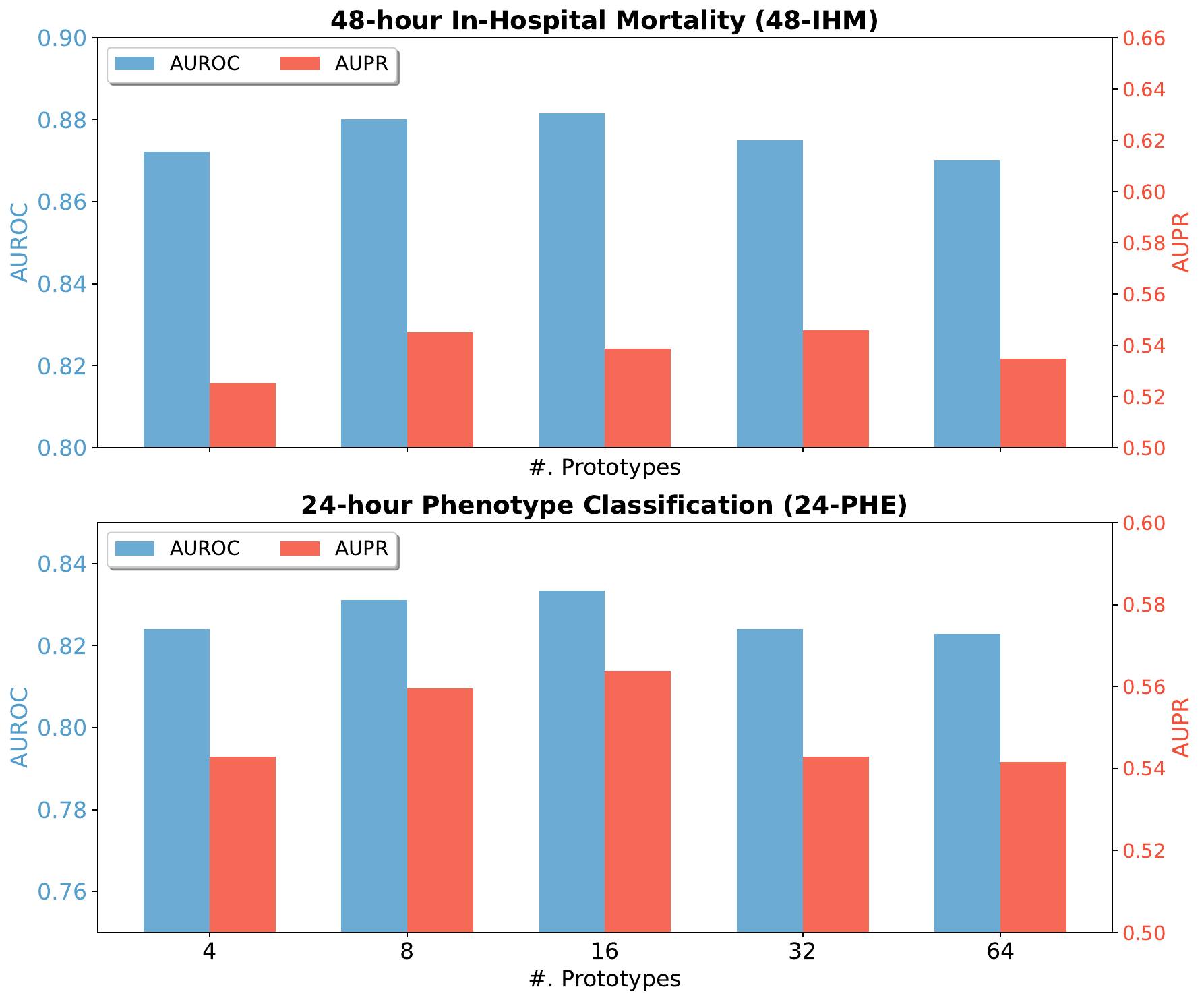}
    \vspace{-0.4cm}
    \label{fig:ablation_number_of_prototypes}
\end{figure}

\para{Ablation Results on Hyperparameters.}
We analyze the effects of the number of prototypes in Fig.~\ref{fig:ablation_number_of_prototypes}.
According to the results, we find that using 16 prototypes achieves the best results, though our model remains robust, with 8 prototypes yielding similar outcomes.
\revise{
Additional experimental  results of hyperparameters and visualization are in Appendix~\ref{sec:ablation_weights} and Appendix~\ref{sec:visualization}.
}

\section{Conclusion}

In this paper, we present the Cross-Modal Temporal Pattern Discovery (\modelname) framework, which captures cross-modal temporal patterns and incorporates them with timestamp-level embeddings for more accurate clinical outcome predictions based on multimodal EHR data. To efficiently optimize the framework, we introduce a  Temporal Pattern Noise Contrastive Estimation (TP-NCE) loss to enhance cross-modal alignment, along with two reconstruction objectives to retain core information from each modality. Our experiments on two clinical prediction tasks using the MIMIC-III dataset demonstrate the effectiveness of \modelname\ in multimodal EHR analysis.
%
\section{Limitations}
\label{sec:limitations}
%

A limitation of our approach is its primary focus on extracting temporal semantics in the embedding space, which affects model interpretability.
%
%
As a potential solution, we plan to explore retrieval-based methods~\cite{patel2024retrieve} or pretrained generative models~\cite{zhao2023rleg} to enhance the interpretability of learned temporal pattern embeddings in future work.
Besides, our work primarily focuses on exploring cross-modal temporal patterns to enhance representation learning, the scenario of missing modalities falls outside the current scope—since shared patterns across modalities cannot be fully captured in their absence.
We plan to thoroughly investigate this direction in future work.
Additionally, our framework is currently designed for specific clinical prediction tasks. In practice, there are various prediction tasks related to multimodal EHR analysis. Extending the proposed method into a more generalized or foundational model capable of handling multiple downstream tasks with minimal training annotations could be more practical and effective.
Our future work will focus on resolving those aspects.

\para{Potential Risks.} The medical dataset used in our framework must be carefully reviewed to mitigate any potential identification risk. Additionally, our framework is developed solely for research purposes and is not intended for commercial use.
Note that AI assistant was used only for polishing the writing of this paper.

\section{Acknowledgement}
This work was supported in part by the Research Grants Council of Hong Kong (27206123, C5055-24G, and T45-401/22-N), the Hong Kong Innovation and Technology Fund (ITS/273/22, ITS/274/22, and GHP/318/22GD), the National Natural Science Foundation of China (No. 62201483), and Guangdong Natural Science Fund (No. 2024A1515011875).

\bibliography{reference}

\clearpage
\appendix

\section*{Appendix}

\begin{table}[ht!]
    \centering
    \caption{
    Dataset description of $48$-IHM and $24$-PHE.
    }
    \label{tab:ihm_statistics}
    \begin{tabular}{c c c c}
    \toprule
    & Training & Validation & Test \\
    \midrule
    \textbf{24-PHE} & $5046$ & $573$ & $1369$ \\
    \midrule
    \textbf{48-IHM} & $4301$ & $466$ & $1183$ \\
    Positive & $644$ & $73$ & $184$ \\
    Negative & $3657$ & $393$ & $999$ \\
    \bottomrule
    \end{tabular}
\end{table}

\begin{table*}[ht!]
    \centering
    \caption{Distribution of 25 phenotypes in 24-PHE task.}
    \label{tab:phe_statistics}
    \resizebox{\linewidth}{!}{
    \begin{tabular}{c c c c c c c c}
    \toprule
    \textbf{Phenotype} & \textbf{Type} & \textbf{Training} & & \textbf{Validation} & & \textbf{Test} & \\ 
    \cmidrule(lr){3-4} \cmidrule(lr){5-6} \cmidrule(lr){7-8}
     &  & \textbf{Positive} & \textbf{Negative} & \textbf{Positive} & \textbf{Negative} & \textbf{Positive} & \textbf{Negative} \\
    \midrule
    Acute and unspecified renal failure & acute & 6066 & 20825 & 2284 & 2284 & 1176 & 4106 \\
    Acute cerebrovascular disease & acute & 2069 & 24822 & 2736 & 2736 & 363 & 4919 \\
    Acute myocardial infarction & acute & 2870 & 24021 & 2632 & 2632 & 588 & 4694 \\
    Cardiac dysrhythmias & mixed & 9011 & 17880 & 1975 & 1975 & 1785 & 3497 \\
    Chronic kidney disease & chronic & 3663 & 23228 & 2564 & 2564 & 711 & 4571 \\
    Chronic obstructive pulmonary disease and bronchiectasis & chronic & 3616 & 23275 & 2542 & 2542 & 685 & 4597 \\
    Complications of surgical procedures or medical care & acute & 5880 & 21011 & 2293 & 2293 & 1202 & 4080 \\
    Conduction disorders & mixed & 1971 & 24920 & 2744 & 2744 & 382 & 4900 \\
    Congestive heart failure; nonhypertensive & mixed & 7623 & 19268 & 2150 & 2150 & 1486 & 3796 \\
    Coronary atherosclerosis and other heart disease & chronic & 8840 & 18051 & 1964 & 1964 & 1787 & 3495 \\
    Diabetes mellitus with complications & mixed & 2612 & 24279 & 2675 & 2675 & 503 & 4779 \\
    Diabetes mellitus without complication & chronic & 5305 & 21586 & 2401 & 2401 & 1032 & 4250 \\
    Disorders of lipid metabolism & chronic & 7841 & 19050 & 2081 & 2081 & 1541 & 3741 \\
    Essential hypertension & chronic & 11340 & 15551 & 1689 & 1689 & 2238 & 3044 \\
    Fluid and electrolyte disorders & acute & 7480 & 19411 & 2107 & 2107 & 1429 & 3853 \\
    Gastrointestinal hemorrhage & acute & 2001 & 24890 & 2746 & 2746 & 427 & 4855 \\
    Hypertension with complications and secondary hypertension & chronic & 3646 & 23245 & 2583 & 2583 & 706 & 4576 \\
    Other liver diseases & mixed & 2485 & 24406 & 2688 & 2688 & 492 & 4790 \\
    Other lower respiratory disease & acute & 1414 & 25477 & 2810 & 2810 & 305 & 4977 \\
    Other upper respiratory disease & acute & 1130 & 25761 & 2814 & 2814 & 238 & 5044 \\
    Pleurisy; pneumothorax; pulmonary collapse & acute & 2516 & 24375 & 2671 & 2671 & 518 & 4764 \\
    Pneumonia (except that caused by tuberculosis or sexually transmitted disease) & acute & 4121 & 22770 & 2502 & 2502 & 769 & 4513 \\
    Respiratory failure; insufficiency; arrest (adult) & acute & 5382 & 21509 & 2365 & 2365 & 1025 & 4257 \\
    Septicemia (except in labor) & acute & 4136 & 22755 & 2494 & 2494 & 793 & 4489 \\
    Shock & acute & 2263 & 24628 & 2700 & 2700 & 456 & 4826 \\
    \bottomrule
    \end{tabular}
    }
\end{table*}

\section{Experimental Setup Details}
\label{sec:dataset_details}
\subsection{More details of Tasks.}

\revise{
We presented the class distributions for 48-IHM and 24-PHE in Table~\ref{tab:ihm_statistics} and Table~\ref{tab:phe_statistics}, respectively.
48-IHM is a binary classification task, whereas 24-PHE is a multi-label classification problem with 25 labels. Notably, our 24-PHE task differs from the phenotype classification problem in the MIMIC-III benchmark but follows the setup in~\cite{zhang2023improving}.
This setup focuses on acute care conditions that arise during ICU stays, where early prediction is critical for timely intervention. To enhance clinical relevance, we use only the first 24 hours of data for phenotype classification rather than the entire admission record. As highlighted in~\cite{yang2021multimodal}, early-stage diagnosis holds greater clinical significance.
}

\revise{
The selection of 25 phenotype labels for the 24-PHE task follows established practices in the MIMIC-III benchmark~\cite{harutyunyan2019multitask}, as also utilized in prior studies like UTDE~\cite{zhang2023improving} and MedFuse~\cite{hayat2022medfuse}.
These labels cover conditions commonly observed in adult ICUs, including 12 critical and life-threatening conditions (e.g., respiratory failure, sepsis), 8 chronic conditions that are often considered comorbidities or risk factors (e.g., diabetes, metabolic disorders), and 5 `mixed' conditions that exhibit characteristics of both chronic and acute conditions.
Phenotype labels were determined using the MIMIC-III ICD-9 diagnosis table. To facilitate the translation and conversion of the abovementioned conditions, we use the Health Cost and Utilization (HCUP) Clinical Classification Software (CCS)\footnote{https://hcup-us.ahrq.gov/toolssoftware/ccs/ccs.jsp}. 
We first mapped each ICD-9 code to its corresponding HCUP CCS category, retaining only 25 categories. 
Diagnoses were then linked to ICU stays using the hospital admission identifier, as ICD-9 codes in MIMIC-III are associated with hospital visits rather than specific ICU stays. 
To reduce label ambiguity, we excluded hospital admissions involving multiple ICU stays, ensuring each diagnosis could be associated with a single ICU stay.
Please find the class distribution of 25 phenotypes in Table~\ref{tab:phe_statistics}.
}

\subsection{Additional Information on Datasets}
%
The 17 variables from the MIMIC-III dataset that we use include 5 categorical variables (capillary refill rate, Glasgow coma scale eye opening, Glasgow coma scale motor response, Glasgow coma scale total, and Glasgow coma scale verbal response) and 12 continuous measures (diastolic blood pressure, fraction of inspired oxygen, glucose, heart rate, height, mean blood pressure, oxygen saturation, respiratory rate, systolic blood pressure, temperature, weight, and pH).

\section{More Details on Baselines}
\label{appendix:more_baseline}

\subsection{Baselines only using MITS}
\begin{itemize}
    \item CNN~\citep{lecun1998gradient}: CNN (Convolutional Neural Network) uses backpropagation to synthesize a complex decision surface that facilitates learning.
    \item RNN~\citep{elman1990finding}: RNN (Residual Neural Network) is trained to process data sequentially so as to model the time dimension of data.
    \item LSTM~\cite{hochreiter1997long}: LSTM (Long short-term memory) is a variant of recurrent neural network. It excels at dealing with the vanishing gradient problem and is relatively insensitive to time gap length.
    \item Transformer~\citep{vaswani2017attention}: Transformer is a powerful deep learning architecture based on attention mechanism. It has great generalisability and has been adopted as foundation model in multiple research domains.
    \item IP-Net~\citep{shukla2019interpolation}: IP-Net (Interpolation-Prediction Network) is a deep learning architecture for supervised learning focusing on processing sparse multivariate time series data that are sampled irregularly.
    \item GRU-D~\citep{che2018recurrent}: GRU-D is based on GRU (Gated Recurrent Unit). It incorporates features of missing data in EHR into the model architecture to improve prediction results.
    
    \item DGM-O~\citep{wu2021dynamic}: DGM-O (Dynamic Gaussian Mixture based Deep Generative Model) is a generative model derived from a dynamic Gaussian mixture distribution. It makes predictions based on incomplete inputs. DGM-O is instantiated with multilayer perceptron (MLP).
    \item mTAND~\citep{shukla2021multi}: mTAND (Multi-Time Attention network) is a deep learning model that learns representations of continuous time values and uses an attention mechanism to generate a consistent representation of a time series based on a varying amount of observations.
    \item SeFT~\citep{horn2020set}: SeFT (Set Functions for Time Series) addresses irregularly-sampled time series. It is based on differentiable set function learning.
    \item UTDE~\citep{zhang2023improving}: UTDE (Unified TDE module) is built upon TDE (Temporal discretization-based embedding). It models asynchronous time series data by combining imputation embeddings and learned interpolation embeddings through a gating mechanism. It also uses a time attention mechanism.
\end{itemize}

\begin{table*}[t!]
    \centering
    \caption{
    Comparison between our method with other baselines on 30-day Readmission task on MIMIC-III.
    We report average performance on three random seeds, with standard deviation as the subscript.
    }
    \resizebox{0.6\textwidth}{!}{
    \begin{tabular}{c | c c c}
        \toprule
        Model & AUROC ($\uparrow$) & AUPR ($\uparrow$)  & F1 ($\uparrow$) \\
        \midrule
        \multicolumn{4}{c}{
        \textit{Methods on EHR.}
        } \\
        \midrule
        CNN~\cite{lecun1998gradient} & 0.757$_\textit{0.003}$ & 0.447$_\textit{0.006}$ & 0.449$_\textit{0.006}$ \\
        RNN~\cite{elman1990finding} & 0.750$_\textit{0.001}$ & 0.449$_\textit{0.008}$ & 0.433$_\textit{0.003}$ \\
        LSTM~\cite{hochreiter1997long} & 0.757$_\textit{0.002}$ & 0.443$_\textit{0.003}$ & 0.430$_\textit{0.015}$ \\
        Transformer~\cite{vaswani2017attention} & 0.749$_\textit{0.006}$ & 0.409$_\textit{0.012}$ & 0.433$_\textit{0.016}$ \\
        DGM-O~\cite{wu2021dynamic} & 0.671$_\textit{0.017}$ & 0.325$_\textit{0.032}$ & 0.382$_\textit{0.020}$ \\
        mTAND~\cite{shukla2019interpolation} & 0.743$_\textit{0.002}$ & 0.437$_\textit{0.001}$ & 0.441$_\textit{0.007}$ \\
        UTDE~\cite{zhang2023improving} & 0.758$_\textit{0.002}$ & 0.453$_\textit{0.004}$ & 0.445$_\textit{0.008}$ \\
        \midrule
        \multicolumn{4}{c}{
        \textit{Methods on Clinical Notes.}
        } \\
        \midrule
        Flat~\cite{deznabi2021predicting} & 0.755$_\textit{0.005}$ & 0.447$_\textit{0.018}$ & 0.437$_\textit{0.015}$ \\
        HierTrans~\cite{pappagari2019hierarchical} & 0.754$_\textit{0.001}$ & 0.425$_\textit{0.005}$ & 0.434$_\textit{0.005}$ \\
        mTAND~\cite{shukla2019interpolation} & 0.757$_\textit{0.001}$ & 0.435$_\textit{0.001}$ & 0.440$_\textit{0.014}$ \\
        \midrule
        \multicolumn{4}{c}{
        \textit{Methods on Multiple modalities.}
        } \\
        \midrule
        MMTM~\cite{joze2020mmtm} & \underline{0.769$_\textit{0.005}$} & \underline{0.469$_\textit{0.006}$} & \underline{0.459$_\textit{0.007}$} \\
        DAFT~\cite{polsterl2021combining} & 0.765$_\textit{0.006}$ & 0.452$_\textit{0.018}$ & 0.458$_\textit{0.005}$ \\
        MedFuse~\cite{hayat2022medfuse} & 0.763$_\textit{0.009}$ & 0.461$_\textit{0.012}$ & 0.454$_\textit{0.013}$ \\
        DrFuse~\cite{yao2024drfuse} & 0.748$_\textit{0.002}$ & 0.443$_\textit{0.014}$ & 0.442$_\textit{0.017}$ \\
        \rowcolor{Gray}
        CTPD (Ours) & \textbf{0.777}$_\textit{\textbf{0.006}}$ & \textbf{0.474}$_\textit{\textbf{0.009}}$ & \textbf{0.461}$_\textit{\textbf{0.019}}$ \\
        \bottomrule
        \end{tabular}
    }
    \label{tab:readm_mimic_iii}
\end{table*}

\subsection{Baselines only using clinical notes}
\begin{itemize}
    \item Flat~\cite{deznabi2021predicting}: Flat encodes clinical notes using a fine-tuned BERT model. It also utilizes an LSTM model that takes in patients’ vital signals so as to jointly model the two modalities. Furthermore, it addresses the temporal irregularity issue of modeling patients' vital signals.
    \item HierTrans~\cite{pappagari2019hierarchical}: HierTrans (Hierarchical Transformers) is built upon BERT model. It achieves an enhanced ability to take in long inputs by first partitioning the inputs into shorter sequences and processing them separately. Then, it propagates each output via a recurrent layer.
    \item T-LSTM~\cite{baytas2017patient}: T-LSTM (Time-Aware LSTM) deals with irregular time intervals in EHRs by learning decomposed cell memory which models elapsed time. The final patient subtyping model uses T-LSTM in an auto-encoder module before doing patient subtyping.
    \item FT-LSTM~\cite{zhang2020time}: FT-LSTM (Flexible Time-aware LSTM Transformer) models the multi-level structure in clinical notes. At the base level, it uses a pre-trained ClinicalBERT model. Then, it merges sequential information and content embedding into a new position-enhanced representation. Then, it uses a time-aware layer that considers the irregularity of time intervals.
    \item GRU-D~\citep{che2018recurrent}: Please refer to the above subsection. 
    \item mTAND~\citep{shukla2021multi}: Please refer to the above subsection.
\end{itemize}

\subsection{Baselines using multimodal EHR}
\begin{itemize}
    \item MMTM~\citep{joze2020mmtm}: MMTM (Multi-modal Transfer Module) is a neural network module that leverages knowledge from various modalities in CNN. It can recalibrate features in each CNN stream via excitation and squeeze operations.
    \item DAFT~\citep{polsterl2021combining}: DAFT (Dynamic Affine Feature Map Transform) is a general-purpose CNN module that alters the feature maps of a convolutional layer with respect to a patient's clinical data.
    \item MedFuse~\citep{hayat2022medfuse}: MedFuse is an LSTM-based fusion module capable of processing both uni-modal and multi-modal input. It treats multi-modal representations of data as a sequence of uni-modal representations. It handles inputs of various lengths via the recurrent inductive bias of LSTM.
    \item DrFuse~\citep{yao2024drfuse}: DrFuse is a fusion module that addresses the issue of missing modality by separating the unique features within each modality and the common ones across modalities. It also adds a disease-wise attention layer for each modality.
\end{itemize}

\begin{table*}[t!]
    \centering
    \caption{
    Comparison between our method with other baselines on 48-IHM and 24-PHE on MIMIC-IV.
    We report average performance on three random seeds, with standard deviation as the subscript.
    The \textbf{Best} and \underline{2nd best} methods under each setup are bold and underlined.
    ``-" denotes the results are close to 0.
    }
    \resizebox{0.9\textwidth}{!}{
    \begin{tabular}{c | c c c |c c c}
        \toprule
        & \multicolumn{3}{c}{48-IHM} & \multicolumn{3}{c}{24-PHE} \\
        Model & AUROC ($\uparrow$) & AUPR ($\uparrow$)  & F1 ($\uparrow$) & AUROC ($\uparrow$) & AUPR ($\uparrow$)  & F1 ($\uparrow$) \\
        \midrule
        \multicolumn{7}{c}{
        \textit{Methods on MITS.}
        } \\
        \midrule
        CNN~\citep{lecun1998gradient} & 74.16$_\textit{0.79}$ & 36.32$_\textit{0.15}$ & 27.34$_\textit{8.89}$ & 67.50$_\textit{0.14}$ & 37.00$_\textit{0.02}$ & 20.90$_\textit{0.81}$ \\
        RNN~\citep{elman1990finding} & 77.40$_\textit{1.10}$ & 38.87$_\textit{0.80}$ & 23.93$_\textit{3.92}$ & 66.89$_\textit{0.49}$ & 36.28$_\textit{0.70}$ & 18.72$_\textit{1.97}$ \\
        LSTM~\cite{hochreiter1997long} & 79.40$_\textit{0.50}$ & 41.95$_\textit{1.09}$ & 34.17$_\textit{1.97}$ & 67.59$_\textit{0.30}$ & 37.21$_\textit{0.49}$ & 20.87$_\textit{1.32}$ \\
        Transformer~\citep{vaswani2017attention} & 75.73$_\textit{2.43}$ & 39.76$_\textit{2.64}$ & 21.45$_\textit{10.75}$ & 67.44$_\textit{0.21}$ & 37.33$_\textit{0.11}$ & 20.44$_\textit{0.40}$ \\
        IP-Net~\citep{shukla2019interpolation} & 77.58$_\textit{0.59}$ & 46.02$_\textit{1.73}$ & 36.8$_\textit{0.93}$ & 68.20$_\textit{0.21}$ & 38.4$_\textit{0.27}$ & 20.62$_\textit{2.58}$ \\
        GRU-D~\citep{che2018recurrent} & 53.72$_\textit{5.31}$ & 17.54$_\textit{2.22}$ & 8.99$_\textit{15.56}$ & 50.53$_\textit{0.60}$ & 23.14$_\textit{0.36}$ & 9.42$_\textit{7.76}$ \\
        DGM-O~\citep{wu2021dynamic} & 70.50$_\textit{10.43}$ & 33.79$_\textit{7.38}$ & 14.14$_\textit{12.42}$ & 59.74$_\textit{2.34}$ & 29.95$_\textit{2.05}$ & 4.33$_\textit{4.12}$ \\
        mTAND~\citep{shukla2021multi} & 80.63$_\textit{0.33}$ & 45.17$_\textit{0.47}$ & 33.01$_\textit{4.57}$ & 66.87$_\textit{0.14}$ & 36.38$_\textit{0.18}$ & 19.07$_\textit{1.28}$ \\
        SeFT~\citep{horn2020set} & 61.97$_\textit{0.96}$ & 25.13$_\textit{0.42}$ & - & 57.10$_\textit{0.04}$ & 26.98$_\textit{0.09}$ & - \\
        UTDE~\citep{zhang2023improving} & 80.89$_\textit{0.10}$ & 45.08$_\textit{0.30}$ & 37.53$_\textit{2.03}$ & 67.68$_\textit{0.05}$ & 37.58$_\textit{0.19}$ & {17.63}$_\textit{0.44}$ \\
        \midrule
        \multicolumn{7}{c}{
        \textit{Methods on CXR Image.}
        } \\
        \midrule
        Flat~\cite{deznabi2021predicting} & 62.37$_\textit{2.42}$ &23.97$_\textit{2.35}$ & 23.38$_\textit{4.34}$ & 66.10$_\textit{0.37}$ & 36.04$_\textit{0.16}$ & 40.40$_\textit{0.54}$ \\
        HierTrans~\cite{pappagari2019hierarchical} & 61.51$_\textit{1.43}$ & 20.96$_\textit{1.62}$ & 17.95$_\textit{15.54}$ & 58.36$_\textit{3.28}$ & 27.84$_\textit{2.16}$ & 33.61$_\textit{5.74}$ \\
        T-LSTM~\cite{baytas2017patient} & 53.62$_\textit{0.88}$ & 16.82$_\textit{0.38}$ & 21.40$_\textit{4.17}$ & 58.56$_\textit{1.45}$ & 28.12$_\textit{1.21}$ & 33.06$_\textit{6.51}$ \\
        FT-LSTM~\cite{zhang2020time} & 48.63$_\textit{4.47}$ & 15.35$_\textit{0.90}$ & 11.85$_\textit{13.75}$ & 54.46$_\textit{3.15}$ & 25.29$_\textit{2.42}$ & 29.12$_\textit{7.55}$ \\
        GRU-D~\citep{che2018recurrent} & 56.02$_\textit{1.05}$ & 18.63$_\textit{0.76}$ & 23.16$_\textit{6.54}$ & 57.94$_\textit{0.39}$ & 27.85$_\textit{0.47}$ & 27.13$_\textit{3.73}$ \\
        mTAND~\citep{shukla2021multi} & 62.80$_\textit{3.42}$ & 24.64$_\textit{4.22}$ & 24.82$_\textit{9.43}$ & 68.31$_\textit{0.68}$ & 38.53$_\textit{0.93}$ & 40.76$_\textit{1.14}$\\
        \midrule
        \multicolumn{7}{c}{
        \textit{Methods on Multiple modalities.}
        } \\
        \midrule
        MMTM~\citep{joze2020mmtm} & 80.65$_\textit{0.79}$ & \underline{48.96$_\textit{0.50}$} & 47.40$_\textit{1.41}$ & 70.28$_\textit{0.44}$ & 39.61$_\textit{0.76}$ & 43.32$_\textit{0.29}$\\
        DAFT~\citep{polsterl2021combining} & \underline{81.87$_\textit{0.12}$} & 47.79$_\textit{0.88}$ & \underline{48.91$_\textit{1.98}$} & \underline{70.87$_\textit{0.24}$} & 40.22$_\textit{0.22}$ & \underline{44.10$_\textit{0.26}$} \\
        MedFuse~\citep{hayat2022medfuse} & 81.60$_\textit{0.28}$ & 48.35$_\textit{1.35}$ & 48.12$_\textit{0.75}$ & 70.82$_\textit{0.37}$ & \underline{40.39$_\textit{0.51}$} & 44.03$_\textit{0.52}$ \\
        DrFuse~\citep{yao2024drfuse} & 80.94$_\textit{0.44}$ & 45.64$_\textit{0.44}$ & 48.58$_\textit{1.35}$ & 70.27$_\textit{0.22}$ & 39.90$_\textit{0.23}$ & 43.43$_\textit{0.09}$\\
        \rowcolor{Gray}
        CTPD \ (Ours) & \textbf{83.53}$_\textit{0.44}$ & \textbf{49.94}$_\textit{0.23}$ & \textbf{49.53}$_\textit{2.39}$ & \textbf{71.96}$_\textit{0.40}$ & \textbf{42.46}$_\textit{0.60}$ & \textbf{44.85}$_\textit{0.61}$ \\
        \bottomrule
        \end{tabular}
    }
    \label{tab:comparison_with_sota_mimic_iv}
    \vspace{-0.5cm}
\end{table*}

\section{More on Experimental Results}
\label{sec:more_experimental_results}

\subsection{Results on 30-day Readmission}
\label{sec:readm}
\revise{
We further evaluated \modelname\ on the 30-day readmission prediction task~\cite{assaf202030} using the MIMIC-III dataset. The results are presented in Table~\ref{tab:readm_mimic_iii}.
This task involves predicting whether a patient will be readmitted based on data from their current ICU stay. Our results show that \modelname\ consistently outperforms baseline methods across all three evaluation metrics, further demonstrating its effectiveness.
}

\subsection{Results on MIMIC-IV}
\label{sec:mimic_iv}
\revise{
To assess the adaptability of our framework to different data sources, we also conducted experiments on the MIMIC-IV dataset. Since MIMIC-IV lacks temporal clinical text, we evaluated our approach using tabular time-series data and chest radiographs as the two input modalities. The results are summarized in Table~\ref{tab:comparison_with_sota_mimic_iv}.
\modelname\ achieved the best performance across all six evaluation settings. Given the relatively small standard deviations, our approach demonstrates statistically significant improvements over previous methods, further validating its generalizability.
}

\begin{table}[t!]
    \centering
    \caption{Ablation results on loss weights. The parameters $\lambda_1$ and $\lambda_2$ control the strength of the $\mathcal{L}_{\text{TPNCE}}$ and $\mathcal{L}_{\text{Recon}}$ losses, respectively.}
    \label{tab:ablation_loss_weights}
    \resizebox{\linewidth}{!}{
    \begin{tabular}{c c | c c c | c c c}
        \toprule
        \multirow{2}{*}{$\lambda_1$} & \multirow{2}{*}{$\lambda_2$} & \multicolumn{3}{c|}{48-IHM} & \multicolumn{3}{c}{24-PHE} \\
        \cmidrule(lr){3-5} \cmidrule(lr){6-8}
        & & AUROC & AUPR & F1 & AUROC & AUPR & F1 \\
        \midrule
        0.1 & 0.1 & 87.21$_{0.36}$ & 53.80$_{0.28}$ & 47.41$_{9.17}$ & 82.93$_{0.05}$ & 55.62$_{0.22}$ & 54.00$_{0.20}$ \\
        0.1 & 0.5 & \textbf{88.15}$_{\mathbf{0.28}}$ & \textbf{53.86}$_{\mathbf{0.65}}$ & \textbf{53.85}$_{\mathbf{0.16}}$ & \textbf{83.34}$_{\mathbf{0.05}}$ & \textbf{56.39}$_{\mathbf{0.17}}$ & \textbf{53.83}$_{\mathbf{0.43}}$ \\
        0.5 & 0.5 & 87.20$_{0.47}$ & 53.77$_{0.47}$ & 42.44$_{7.45}$ & 82.59$_{0.09}$ & 55.22$_{0.04}$ & 53.42$_{0.04}$ \\
        1.0 & 0.5 & 86.59$_{2.19}$ & 51.93$_{5.31}$ & 47.93$_{3.13}$ & 82.44$_{0.03}$ & 54.84$_{0.18}$ & 53.29$_{0.29}$ \\
        1.0 & 1.0 & 86.64$_{2.10}$ & 52.34$_{5.79}$ & 50.09$_{4.34}$ & 82.54$_{0.48}$ & 55.03$_{0.39}$ & 53.94$_{1.31}$ \\
        1.0 & 2.0 & 85.56$_{1.19}$ & 50.66$_{4.17}$ & 44.67$_{3.88}$ & 76.88$_{1.84}$ & 41.89$_{3.05}$ & 44.83$_{2.31}$ \\
        \bottomrule
    \end{tabular}
    }
\end{table}

\begin{figure}[ht!] 
  \begin{minipage}[b]{0.43\linewidth}
    \centering
    \includegraphics[width=\linewidth]{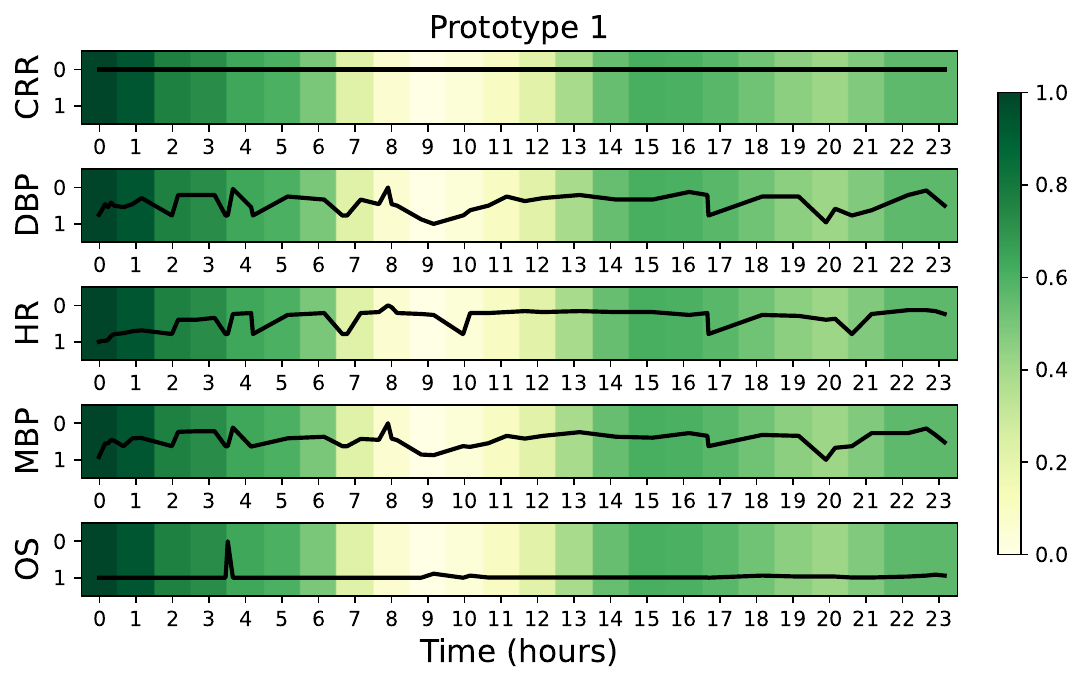} 
  \end{minipage}
  \hfill
  \begin{minipage}[b]{0.43\linewidth}
    \centering
    \includegraphics[width=\linewidth]{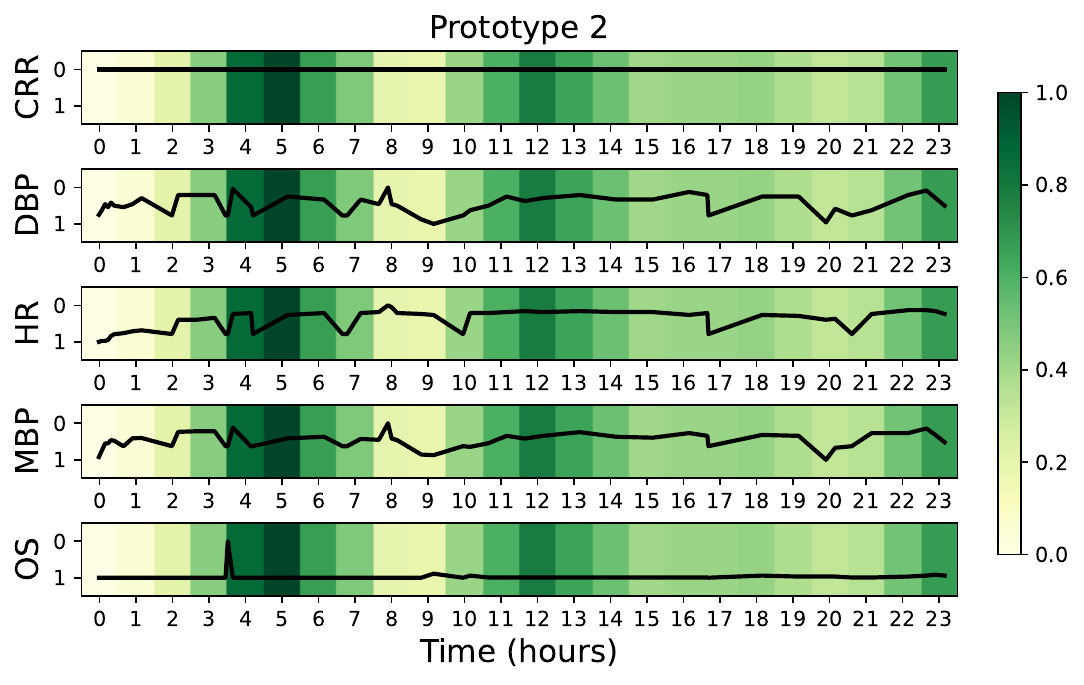} 
  \end{minipage} 
  \hfill
  \begin{minipage}[b]{0.43\linewidth}
    \centering
    \includegraphics[width=\linewidth]{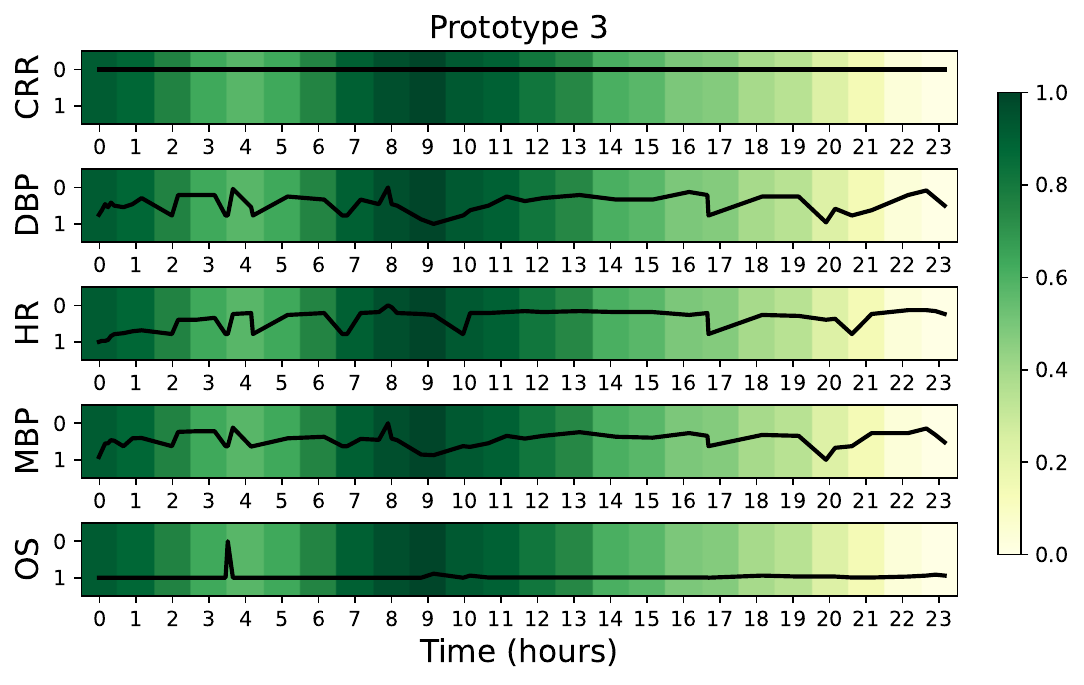} 
  \end{minipage}
  \hfill
  \begin{minipage}[b]{0.43\linewidth}
    \centering
    \includegraphics[width=\linewidth]{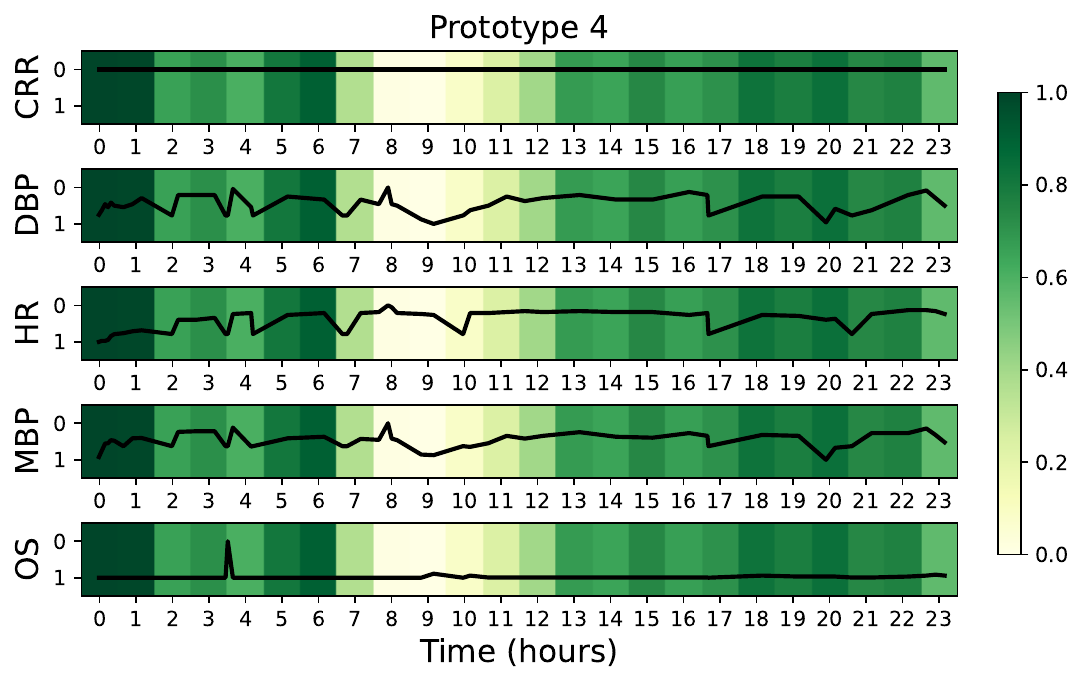} 
  \end{minipage} 
  \caption{
    \revise{
    Visualization of the learned prototypes in our \modelname \ framework.
    Here we select $5$ representative clinical variables to visualize the time series.
    `CPR" denotes ``Capillary Refill Rate", ``DBP" denotes ``Diastolic blood pressure", `, ``HR" denotes ``Heart Rate", ``MBP" denotes ``Mean blood pressure" and ``OS" denotes ``Oxygen saturation".
    }
  }
  \label{fig:attention_map}
  \vspace{-0.5cm}
\end{figure}

\subsection{Ablation Results on Loss Weights}
\label{sec:ablation_weights}
We analyze the effects of loss weights $\lambda_1$ and $\lambda_2$, shown in Table~\ref{tab:ablation_loss_weights}.
For the loss weights, $\lambda_1 = 0.1$ and $\lambda_2 = 0.5$ consistently yield the best performance across all settings. 
In contrast, larger values (rows 4–6) lead to performance drops, likely due to the model overemphasizing alignment and reconstruction losses at the expense of prediction accuracy.
Moderate changes to $\lambda_1$ and $\lambda_2$ (rows 1–3) yield stable results, particularly on the complex 24-PHE task, indicating robustness. However, the simpler 48-IHM task is more sensitive, possibly due to its limited temporal complexity and tendency to overfit.
In practice, we adopt $\lambda_1 = 0.1$ and $\lambda_2 = 0.5$ to maintain a balanced loss scale without extensive tuning. Future work may explore finer-grained searches to better understand the trade-offs across tasks.

\subsection{Visualization of Assignment Weights of Prototypes.}
\label{sec:visualization}
\revise{
Fig.~\ref{fig:attention_map} presents the distribution of assignment weights across different time scales for a 48-IHM example.
With time window sizes denoted by $T$, our model utilizes three time scales with $20$, $10$, and $5$ prototypes respectively. 
The variation in assignment weights across these scales, as showcased in the figure, underlines our model’s proficiency in capturing and differentiating temporal patterns at varying scales.
We will try to interpret these learned temporal pattern embeddings in our future work.
}

\end{document}